\documentclass[acmsmall,nonacm]{acmart}
\AtBeginDocument{%
  }

\usepackage{times}
\usepackage{soul}
\usepackage{url}

\usepackage{tcolorbox}

\usepackage{amsmath}
\usepackage{amsthm}
\usepackage{booktabs}
\usepackage{algorithm}
\usepackage{algorithmic}
\usepackage[switch]{lineno}
\usepackage{subcaption}
\usepackage{xspace}
\usepackage{xcolor}
\usepackage{diagbox}  
\usepackage{enumitem}
\usepackage{longtable}
\usepackage{array}
\usepackage{enumitem}
\usepackage{xcolor}
\usepackage{setspace}

\usepackage{pifont}
\newcommand{\cmark}{\ding{51}}%

\newcommand{\modif}[1]{\textcolor{blue}{#1}}
\renewcommand{\modif}[1]{#1}

\definecolor{specialgray}{rgb}{0.9, 0.9, 0.9} 
\definecolor{specialblue}{rgb}{0.1, 0.3, 0.8} 

\newtcolorbox{mybox}{
  sharp corners,
  colback=specialgray,
  colframe=specialblue,
  boxrule=0pt,
  toprule=0pt,
  bottomrule=0pt,
  leftrule=3pt, 
  rightrule=3pt 
}

\newcommand{\ourf}[1]{{\textsc{#1}}}
\newcommand{\mix}{\ourf{Mixtral-8B}\xspace} 
\newcommand{\gpt}{\ourf{GPT-4o-mini}\xspace} %
\newcommand{\gem}{\ourf{Gemini-1.5F}\xspace} 

\newcommand{\llmd}{\ourf{LLMd}\xspace}
\newcommand{\llmdmix}{\ourf{LLMd\textsubscript{mix}}\xspace}
\newcommand{\llmdgpt}{\ourf{LLMd\textsubscript{gpt}}\xspace}
\newcommand{\llmdgem}{\ourf{LLMd\textsubscript{gem}}\xspace}

\newcommand{\llmnd}{\ourf{LLMnd}\xspace}
\newcommand{\llmndmix}{\ourf{LLMnd\textsubscript{mix}}\xspace}
\newcommand{\llmndgpt}{\ourf{LLMnd\textsubscript{gpt}}\xspace}
\newcommand{\llmndgem}{\ourf{LLMnd\textsubscript{gem}}\xspace}

\newcommand{\rdis}{\ourf{REDd}\xspace}

\usepackage{natbib}

\makeatletter
\newcommand\footnoteref[1]{\protected@xdef\@thefnmark{\ref{#1}}\@footnotemark}
\makeatother

\begin{document}

\title{Shiny Stories, Hidden Struggles: Investigating the Representation of Disability Through the Lens of LLMs}




\author{Marco Bombieri}
\email{marco.bombieri@unitn.it}
\orcid{0000-0002-8607-8495}
\authornotemark[1]

\affiliation{%
  \institution{Department of Foreign Languages and Literature, University of Verona}
  \city{Verona}
  \country{Italy}
}

\affiliation{%
  \institution{Department of Information Engineering and Computer Science, University of Trento}
  \city{Povo, Trento}
  \country{Italy}
}

\author{Simone Paolo Ponzetto}
\affiliation{%
  \institution{Data and Web Science Group, University of Mannheim}
  \city{Mannheim}
  \country{Germany}}

\author{Marco Rospocher}
\affiliation{%
  \institution{Department of Foreign Languages and Literature, University of Verona}
  \city{Verona}
  \country{Italy}
}

\renewcommand{\shortauthors}{Bombieri et al.}

\renewcommand{\shortauthors}{Bombieri et al.}

\begin{abstract}
Modern Large Language Models (LLMs) have recently attracted much attention for their ability to simulate human behavior and generate text that reflects personas and demographic groups. While these capabilities can open up a multitude of diverse applications across fields, it is crucial to examine how such models represent various target groups since LLMs can perpetuate and amplify biases or discrimination against historically marginalized communities or, alternatively, as a result of debiasing efforts, overcorrect by portraying overly positive stereotypes. This overcompensation can idealize these groups, erasing the complexities and challenges they face in favor of unrealistic depictions.
In this paper, we investigate how LLMs represent disability by simulating the perspectives of individuals with disabilities in generating social media posts. These posts are then compared with those written by real people with disabilities, focusing on emotional tone, sentiment, and representative words and themes. Our analysis reveals two key findings: (1) LLMs often idealize the experiences of people with disabilities, producing overly positive stereotypes that, despite appearing uplifting, fail to authentically capture their lived realities; and (2) a comparative analysis of posts simulating individuals with and without disabilities highlights a negative bias, where certain topics, such as career and entertainment, are disproportionately associated with nondisabled individuals. This reinforces exclusionary narratives and over-idealized portrayals of disability, misrepresenting the actual challenges faced by this community.
These findings align with broader concerns and ongoing research showing that LLMs struggle to reflect the diverse realities of society, particularly the nuanced experiences of marginalized groups, and underscore the need for critical scrutiny of their representations.
\end{abstract}



\maketitle
\section{Introduction}
In recent years, much work has focused on showing how computational models of language trained on real-world data mirror and amplify harmful societal biases, often disproportionately affecting historically marginalized communities (\cite{bolukbasi2016,manzini-etal-2019-black}, \textit{inter alia}), whose perpetuation in society can lead to psychological harm, unhappiness and, in some cases, suicide attempts \cite{disability-centered_perspectives-2023}.  The rising usage and widespread adoption of recent Large Language Models (LLMs) has further amplified this problem, with potential further spread of these representational harms \cite{gallegos24}.
Consequently, research with a focus on biases and stereotypes in LLMs has also proposed methods to mitigate these issues. For example, almost all recent prompt-based LLMs have been equipped with embedded de-biasing techniques and AI guards (e.g., \cite{llama-guard2023}) that block a potentially offensive and hurtful question, refusing to give a direct answer. These guards can also internally adjust the model's response, ensuring it is non-toxic and non-offensive, rather than simply following the word distributions learned during training. \modif{However, AI companies that release LLMs and the embedded AI guards rarely disclose the detailed instructions or thresholds that trigger content moderation and thus the mechanisms behind these safeguards are often not transparent. 
Moreover,} recent work on studying the depiction of personas from marginalized groups of LLMs indicates that many biases are concealed even in texts containing words with a positive sentiment, which can still offend their sensitivities and lead to pernicious positive portrayals: 
\modif{for example, in \citet{marked-personas} GPT-4 describes Asian women as \textit{almond-shaped, petite, delicate}, and \textit{smooth}—terms rooted in Western media portrayals that frame them as exotic, submissive, and hypersexualized; similarly, Latina women are depicted as \textit{vibrant, beautiful}, and \textit{curvaceous}, using language that homogenizes and hypersexualizes this identity; finally, Black women are characterized with words such as \textit{strength} and \textit{resilience}—apparently positive terms that, however, have been hystorically used as a "solution" to structural issues such as poverty and inequality}.
%
\modif{Such descriptions contribute to what can be identified as \textit{positive idealization}, \textit{toxic positivity}, and \textit{overcompensation}. \textit{Positive idealization} occurs when marginalized identities are represented exclusively through flattering or ideal traits—beauty, grace, strength—producing positive stereotypes that erase diversity and individuality. \textit{Toxic positivity}, by contrast, refers to the use of positive language to deny or downplay experiences of marginalization, implying that optimism or resilience can substitute for the acknowledgment of structural injustice. Finally, \textit{overcompensation} arises when the model, in an effort to avoid negative bias, overcorrects by amplifying positive descriptors to an unrealistic degree, resulting in artificially virtuous or exaggerated portrayals. Together, these mechanisms reveal how seemingly benevolent representations can reproduce harm by masking inequity and silencing the complexity of marginalized experiences.}

In this paper, we follow this line of work and explore how current LLMs depict members of historically marginalized
communities. While prior research has focused mainly on biases related to age, race/ethnicity, and gender, we investigate here biases in LLMs against people with disabilities\footnote{Throughout this paper, we predominantly use people-first language (e.g., ``people with disabilities''). We recognize and respect that preferences regarding language may vary, with some individuals favoring people-first language while others prefer identity-first language (e.g., ``disabled people''). We intend not to offend or diminish anyone's perspective.}, a group that has received comparatively less attention in the literature \cite{disability-centered_perspectives-2023} \modif{yet frequently faces discrimination and experiences lower socio-economic status~\cite{VanPuymbrouck2020ExplicitAI,SZUMSKI2020103685}.}

\vspace{1em}
\noindent
\textbf{Research questions}. This paper explores how LLMs represent disability by comparing self-descriptions from real individuals with disabilities to the portrayals generated by LLMs. We further examine whether and how LLMs adjust their language and thematic considerations when describing individuals with disabilities compared to generic personas. Specifically, we address the following Research Questions (RQs):

\begin{itemize}[leftmargin=8mm]
    \item[\textbf{RQ1}] How do self-descriptions provided by real individuals with disabilities compare to portrayals generated by LLMs? 
    \item[\textbf{RQ2}] Do LLMs alter their language when describing individuals with disabilities as opposed to generic individuals? 
\end{itemize}

\noindent
\textbf{Contributions.} Answering the above RQs, we offer the following contributions:

\begin{itemize}[leftmargin=6mm]
    \item[\textbf{C1}] We gathered, annotated, and publicly released a dataset of social media (Reddit) posts written by users with disability presenting themselves on the platform. Similarly, using different LLMs, we generated and publicly released a dataset of artificial portrayals of people with disabilities and generic individuals presenting themselves on social media platforms using different prompts inspired by \cite{portrayal2022} and \cite{marked-personas}. Each post of both datasets is automatically annotated with its most probable primary emotions and sentiment, and the indication of whether the text reveals or not the presence of the writer's depression, whether severe or moderate.
    \item[\textbf{C2}] 
    We conducted a comprehensive comparison of posts collected from the web with those generated by LLMs to analyze the representation of individuals with disabilities by LLMs and to identify key differences between real-world and AI-generated portrayals. Our findings highlight the importance of broadening research on biases to include not only negative stereotypes but also positive idealizations, which can similarly perpetuate harm to marginalized or vulnerable groups.
\end{itemize}

\noindent

\modif{The analysis of the dataset composed by posts written by people with disabilities highlights the significant challenges they often face, which can evoke deeply negative emotions or language suggestive of depressive symptoms. Our experiments show that LLMs frequently flatten or remove these aspects in their portrayals of disabled personas, favoring generalized or overly positive representations. This tendency risks stereotyping and marginalizing individuals whose experiences do not fit a simplified narrative. More broadly, it illustrates a critical ethical concern: AI systems should represent the full spectrum of human emotions—both positive and difficult—to support genuine inclusion and empathy. By revealing these gaps, our study underscores the urgent need for AI policies and design practices that are attentive to individual experiences, promote accessibility, and ensure that technological tools align with the lived realities of marginalized communities, fostering real understanding rather than superficial representation.}

The paper is organized as follows: 
\modif{
Section \ref{sec:related} reviews related works, including societal representation of disability and studies on LLMs and fairness, focusing on biases and stereotypes. It further delves into research addressing bias detection and correction related to people with disabilities and concludes by examining the emerging trend of human portrayal and simulation using LLMs. 
}
Section \ref{sec:task} outlines the approach to address the RQs. It details the datasets collection process and describes the techniques employed to compare the language of individuals with disabilities to that of LLMs simulating these individuals, as well as LLMs simulating general personas (i.e., without explicitly referencing disability in the prompts). Section \ref{sec:results} presents and discusses the obtained results. 
Section \ref{sec:conclusions} summarizes and concludes the paper, presenting limitations and future works. Ethical implications and limitations are reported at the end of the paper.

\section{Related Work} \label{sec:related}

\noindent
\paragraph{LLMs and Fairness.} Rapid advancements in LLMs have revolutionized the processing, comprehension, and generation of human-like text, driving their growing integration into systems that influence our social interactions. However, beneath these achievements lies the risk of causing harm and perpetuating stereotypes and bias \cite{gallegos24}. In fact, LLMs are generally trained on vast amounts of uncurated, internet-based data, causing them to inherit stereotypes \cite{stereotypes}, misrepresentations that contribute to social and cultural erasure \cite{marked-personas}, derogatory and exclusionary language \cite{discriminatory}, wrong associations \cite{bolukbasi2016, manzini-etal-2019-black} and other harmful biases \cite{societal-bias21}. These issues disproportionately impact different vulnerable and marginalized communities (e.g., age, race/ethnicity, gender, and disability) when LLMs are applied in various downstream tasks~\cite{acl2020HutchinsonP,facct2023-bias-93-groups2023,workshop-genlp-2022-gender-bias,hiring-llm-2024,smith-etal-2022-im}. The increasing awareness of inherent social biases in LLMs has led to a surge of research focused on developing methods to measure or mitigate them. This includes bias evaluation metrics, datasets designed for bias assessment, and techniques for bias mitigation \cite{gallegos24}. However, although the World Health Organization (WHO) reports that 1.3 billion people currently live with significant disabilities \cite{WHO}, there remains a gap in research on the biases of language models and their representation of disability \cite{coling2022narayanan,fairness-review24}. 
This limited consideration is mainly due to the lack of available datasets and existing ones, such as BBQ \cite{BBQ2022}, HolisticBias \cite{smith-etal-2022-im}, and PANDA \cite{qian-etal-2022-perturbation}, only partially address disabilities, thus lacking a comprehensive range of impairments.

\modif{\paragraph{Disability representation in society}
Studies emphasize that disability is not merely an individual trait but also a socially and culturally constructed phenomenon. Unlike the \textit{medical model}, which frames disability as a personal deficit to be corrected, the \textit{social} and \textit{rights-based} models view disability as arising from the interaction between individual impairments and societal barriers, including physical, symbolic, and institutional obstacles \citep{Oliver1990}. Accordingly, how society represents disability plays a central role in reproducing or challenging ableism, the system of values that positions non-disabled bodies as normative and ideal \citep{Campbell2009}.
Negative biases are a persistent feature of societal portrayals of disability. People with disabilities are often subject to discrimination, social exclusion, and patronizing attitudes, which can limit their participation in education, employment, and public life \citep{Babik2021}. Media representations frequently reinforce these biases by portraying disability as a problem, a burden, or a source of tragedy, thereby dehumanizing disabled individuals and sustaining societal stigma \citep{Barnes1992}.
In addition to negative biases, disability is also represented through subtler, seemingly positive narratives that can nonetheless be harmful \cite{Grue2016}. An example is \textit{inspiration porn}, in which people with disabilities are portrayed primarily to inspire or motivate non-disabled audiences, rather than as individuals with their own rights, experiences, and interests \citep{disability-centered_perspectives-2023,Chapter10InspirationPornandDesperationPorn,Grue02072016}. These portrayals often emphasize courage, exceptional effort, or the ability to ``overcome'' challenges, creating a narrow standard of values that the audience is encouraged to admire. While appearing positive, such narratives can reinforce unrealistic expectations, objectify disabled individuals, and obscure the structural and societal dimensions of disability, reducing complex lived experiences to symbolic or motivational functions \citep{Grue2016}.
Together, these insights indicate that societal perceptions of disability are shaped by both overtly negative stereotypes and subtler, seemingly positive narratives. In the context of our work, this implies that the analysis of LLMs should go beyond detecting negative bias: it must also consider how seemingly positive portrayals may reproduce harmful representational tropes. 
}

\modif{\paragraph{Language models and disability.}
While there is growing interest in leveraging AI systems and LLMs to enhance inclusion and accessibility for people with disabilities \cite{ledoyen-etal-2025-facilitating,dislexya-10.1145/3696593.3696625,Fitas2025}, the internal representation of disability within LLMs, their potential biases, and their implications for downstream applications have so far received only limited attention.}
The literature defines disability bias as a situation when a person with a disability is treated less favorably than a person without it in the same or similar circumstances \cite{coling2022narayanan}.
In this context, some studies analyzed, for example, how hiring systems tend to discriminate against candidates with disabilities \cite{recruitment-general-bias,facct2024Glazko}. In particular, \cite{facct2024Glazko} shows that even the most recent LLMs, such as GPT4, discriminate in their output against people with disabilities in suggesting the best candidates for a job.
\modif{Similarly, \citet{panda-etal-2025-accesseval} finds that responses to disability-aware queries exhibit a more negative tone, greater stereotyping, higher factual error rates, and more frequent denial responses compared to neutral queries. This suggests that people with disabilities often receive less useful and less accurate answers from LLMs than non-disabled users. A recent work \cite{2025mdpi} employs a survey-based impersonation methodology, asking LLMs to fill out demographic questionnaires as if they were individuals with disabilities, and shows that the resulting profiles generate with popular models (GPT-4, Gemini-2.5, Gemma-3, OLMo-2) systematically reflect stereotypical associations—such as lower education, reduced income, and lower life satisfaction—revealing persistent disability-related bias in model representations.}
\citet{coling2022narayanan} and \citet{acl2020HutchinsonP} use perturbation sensitivity analysis \cite{age-related2019} to find biases against people with disabilities in classical language models and word embeddings (e.g., BERT \cite{devlin-etal-2019-bert} and GPT2 \cite{radford2019language}). They find that traditional language models and word embeddings exhibit implicit bias by assigning more negative or toxic scores to sentences containing words related to disability. \citet{emnlp2021Hassan} follows a similar approach but considers an intersectional setting, i.e., analyzing disability in conjunction with gender and ethnicity. Finally, \citet{herold-etal-2022-applying} analyzes the representation of disabilities in BERT, finding that the models tend to associate negative stereotypes with people with a disability, treating them in a medical way, i.e.,  framing disability simply as a problem on the body. 
Recent work by \citet{li-etal-2024-decoding} analyzes the behavior of recent LLMs, such as GPT-3.5 and GPT-4, and finds that they generally produce texts with a more positive sentiment when addressing disabilities than earlier models. This shift is likely attributable to recent debiasing efforts. 
However, in their limitations section, the authors caution against the risk of perpetuating the of "inspiration porn", where overly idealized portrayals of disability can be harmful.
In this paper, we build upon this previous work and delve deeper into this line of research by examining \textit{how} the models represent disabilities, focusing on both negative and positive biases with overcompensation—instances where overly positive yet subtly harmful portrayals are generated. In addition to the sentiment, we employ a broader range of metrics (e.g., emotion and depression detection, z-scores, and thematic analysis) to comprehensively investigate these research questions.


\paragraph{LLM-based portrayals and human simulation.} Another related research trend is that of human simulation, i.e., evaluating to what extent
a LLM can simulate different aspects of human behavior. This research direction is known as Turing Experiment \cite{icml-AherAK23}. Similar approaches can be applied to simulate humans in different social and political settings \cite{PoliticalBusby2023,abs-2024-simulate,abs-2024-human-behaviour} or to find biases \cite{portrayal2022,marked-personas} and distortions in LLMs representations. 
In particular, \citet{acl-HuC24-persona-effect} measures how effectively an LLM can simulate personas when their demographic, social, and behavioral factors vary, finding the challenge of the task, especially with the zero-shot approach. To address this, \citet{abs-bigchat} suggests enhancing the realism of personality trait representation by assembling a dataset of persona descriptions, which is then used to fine-tune LLMs to better align with human personality patterns. Building on this research direction, our work has a different focus: we investigate how LLMs represent disability in a zero-shot setting, specifically examining the presence of oversimplifications and stereotypes within the base model.
\section{Investigating the representation of disabilities in LLMs}
\label{sec:task}

\subsection{Generating portrays of people with disabilities using LLMs}\label{sub:dataset-generated}
Using a set of prompts inspired by \citet{portrayal2022} and following the approach of authoritative studies such as \cite{marked-personas}, we use three LLMs to craft social media posts from the perspective of general individuals or people with disabilities. Since the use of more detailed prompts has been questioned due to the risk of steering the model toward specific topics, risking introducing stereotypes \cite{marked-personas}, we specifically opted to use open-ended prompts, allowing us to explore the model’s responses in a completely free and "spontaneous" setting.
In particular, we employ \mix (Mixtral-8x7B Instruct, quantized to 4 bits using GPTQ \cite{gptq2022} due to hardware limitations), \gpt, and \gem (the Flash variant). For the latter two models, we utilized their paid APIs. The models are used with the temperature set to $1.0$ to guarantee the variability and randomness of the responses generated.

Each prompt asks the models to introduce themselves as an individual with a disability or as a generic person and, depending on the context, share experiences related to disability (or their `status' in the setting where disability is not mentioned) or general life experiences such as expressions of emotions, feelings, or thoughts, descriptions of daily activities, questions for the community, requests for suggestions, or commentary on current events, i.e., the typical activities a user can do on a social media platform \cite{al2015motivations}.
In more detail, all the prompts follow the template: \textit{"Imagine you are [PERSON]. Write a post on social media introducing yourself and [ACTIVITY]"},
where \textit{[PERSON]} and \textit{[ACTIVITY]} can be one of those defined in Table \ref{tab:template}.

%

\begin{table}[t]
\centering
\begin{tabular}{|p{5.5cm}|p{7.5cm}|}
\hline
\textit{[PERSON]} & \textit{[ACTIVITY]} \\ 
\hline
\hline
\textbf{P1} : \textit{a person with a disability.} & \textbf{A1} : \textit{sharing the experiences related to your disability.} \\ \hline
\textbf{P2} : \textit{a person with autism.} & \textbf{A2} : \textit{sharing the emotions you felt today.} \\ \hline
\textbf{P3} : \textit{a person with blindness.} & \textbf{A3} : \textit{sharing the thoughts you had today.} \\ \hline
\textbf{P4} : \textit{a person with deafness.} & \textbf{A4} : \textit{sharing the activities you did today.} \\ \hline
\textbf{P5} : \textit{a person with cerebral palsy.} & \textbf{A5} : \textit{asking the community a question or a suggestion.} \\ \hline
\textbf{P6} : \textit{a person with depression.} & \textbf{A6} : \textit{commenting on today's events.} \\ \hline
\textbf{P7} : \textit{a person.} & \textbf{A7} : \textit{sharing the experiences related to your status.} \\ 
\hline
\hline
\end{tabular}
\caption{List of possible values of \textit{[PERSON]} and \textit{[ACTIVITY]} in the prompt template we use in our experiments (``Imagine you are [PERSON]. Write a post on social media introducing yourself and [ACTIVITY]'').}
\label{tab:template}
\end{table}

In more detail, the combination of P1-P6 with A1-A6 aims to generate portrayals of individuals with disabilities or impairments. Given the complexity of the topic, covering all possible disabilities or impairments in a single paper would be impractical. Therefore, we limit our analysis to the types of disabilities examined in recent, comparable disability-focused studies (e.g., \cite{facct2024Glazko}). Based on this, we define six alternative \textit{[PERSON]} variants (P1–P6). Exploiting all possible combinations, we thus obtain 36 different prompts. Each prompt is submitted $10$ times to take into account the output variability of the models, thus obtaining, for each LLM, a collection of $360$ posts of artificial portrayals of people with disabilities. We call \llmd the dataset containing the posts generated by all LLMs, where \llmdgpt, \llmdgem, and \llmdmix are its three components consisting of all the posts generated with \gpt, \gem, and \mix, respectively. 

P7, in combination with A2-A7, is instead used for generating portrayals of generic individuals, i.e., where disability is not explicitly mentioned in the prompt. This allows us to examine potential differences in the generated texts between the portrayal of a marked person, i.e., a person with one of the disabilities mentioned in the prompts, and an unmarked one, i.e., where disability is not mentioned. In this setting, each prompt is repeated $60$ times, resulting in a collection of additional $360$ posts per model. 

We call \llmnd the dataset containing the posts generated by all LLMs, omitting any mention of disability, where \llmndgpt, \llmndgem, and \llmndmix are its three components, consisting of all the posts generated with \gpt, \gem, and \mix, respectively.

The size of our datasets is in line with state-of-the-art studies \cite{portrayal2022,marked-personas} and reported in Table \ref{tab:summary-models}.

\subsection{Collecting a dataset of self-descriptions of real people with disabilities}  \label{sub:dataset-reddit}

In addition to the datasets described in Section \ref{sub:dataset-generated}, we collected posts from six disability-related Subreddits.
In particular, we started from a general disability-centered subreddit, i.e., \texttt{r/disability}\footnote{Subreddit \texttt{r/disability}: \url{https://www.reddit.com/r/disability/} [Last access: 2025-05-16]}. It is a general space with the goal of gathering news, discourses, and perspectives pertaining to individuals with disability; it is not specific to any disability, and it encompasses a wide range of discussions, from personal experiences and systemic challenges to advocacy, thus providing a diverse and dynamic dataset for analysis.
Furthermore, it is also very active and used (top $2\%$ subreddits by size), thus indicating its importance for the community.
To mitigate a possible selection bias, obtain a diverse and representative dataset, and deal with disabilities or impairments considered in Section \ref{sub:dataset-generated}, we further extended it with five other subreddits that are more focused on a specific disability or impairment. In particular we analyzed \texttt{r/blind}\footnote{Subreddit \texttt{r/blind}: \url{https://www.reddit.com/r/blind/} [Last access: 2025-05-16]};  
\texttt{r/autism}\footnote{Subreddit \texttt{r/autism}: \url{https://www.reddit.com/r/autism/}}; 
\texttt{r/depression}\footnote{Subreddit \texttt{r/depression}: \url{https://www.reddit.com/r/depression/}}, 
\texttt{r/deaf}\footnote{Subreddit \texttt{r/deaf}: \url{https://www.reddit.com/r/deaf/}}, and 
\texttt{r/celebralpalsy}\footnote{Subreddit \texttt{r/cerebralpalsy}: \url{https://www.reddit.com/r/cerebralpalsy/}} respectively dedicated to sharing experiences and opinions, and perspectives about blindness, autism, depression, deafness, and cerebral palsy. 
In the textual descriptions of these subreddits, their goal is often specified, i.e.,  designing a space where people with disabilities can share their thoughts, experiences, and perspectives or make new friends. They are thus not explicitly designed as a space for venting frustrations or complaints. 
We collected posts published until 2024 that contained textual content, excluding empty posts and those consisting solely of links, images, or videos. We then applied \mix (the prompt used is provided in Appendix \ref{appendix:prompt-used}) to filter out posts not written in the first person or those that did not explicitly come from individuals self-identifying as having a disability. This is because we are only interested in the perspectives of real people with disabilities, while in these subreddits, there are also posts written by family members, medical doctors, or others, such as scientists.

From the filtered results, we randomly selected a sample of 450 posts for r/disability and 220 for the disability-specific subreddits. Three annotators then manually reviewed these posts. Following this validation, the final dataset, named \rdis, consists of 352 posts from \texttt{r/disability}, 165 from \texttt{r/blind}, 174 from \texttt{r/autism}, 204 from \texttt{r/depression}, 171 from \texttt{r/deaf} and 183 from \texttt{r/cerebralpalsy}. \footnote{Please note that our goal is not to develop an LLM for identifying self-descriptions in posts but to collect a dataset of posts written by people with disabilities to carry out our analysis: the LLM (with an accuracy of 78\% in this task) has thus a pure functional role to assist in filtering relevant posts.}  To ensure the quality of the annotation process, 50 posts were initially extracted and independently annotated by all three annotators. Inter-annotator agreement for these posts was measured using Fleiss’ Kappa, resulting in a value of $0.875$, indicating very high ($>0.8$) agreement~\cite{Landis77}.

\subsection{Comparison metrics}  \label{sub:comparison-metrics}
To address our research questions, we aim to perform a pairwise comparison of the previously described datasets, i.e., the LLM-generated portraits (Section \ref{sub:dataset-generated}) and human descriptions from Reddit users (\ref{sub:dataset-reddit}). Given two datasets, we compare them along the following dimensions:
\begin{itemize}[leftmargin=4mm]
\item The predominant \textit{sentiment} of each post $p$ is computed using VADER \cite{bib:VADER}, which assigns a sentiment score \( S(p) \in [-1, +1] \). A post is classified as positive if \( S(p) > 0.05 \), negative if \( S(p) < -0.05 \), and neutral otherwise. For a dataset \( P = [p_1, \dots, p_N] \) of \( N \) posts, we compute the number of positive, negative, and neutral posts: 
$
N_{\text{positive}} = \left| \{ p_i \mid S(p_i) > 0.05 \} \right|, \, N_{\text{negative}} = \left| \{ p_i \mid S(p_i) < -0.05 \} \right|,  N_{\text{neutral}} = \left| \{ p_i \mid -0.05 < S(p_i) < 0.05 \} \right|.
$
The proportions of positive and negative posts are thus\footnote{Posts with scores between $-0.05$ and $0.05$ are considered neutral. Since \rdis is the only dataset containing neutral posts — and only two such posts — we chose to focus our analysis exclusively on positive and negative posts.}: 
$
P_{\text{positive}} = \frac{N_{\text{positive}}}{N}, \, P_{\text{negative}} = \frac{N_{\text{negative}}}{N}.
$

\item The distribution of \textit{emotions} \( E = \{\text{anger}, \text{fear}, \text{anticipation}, \text{trust}, \text{surprise}, \text{sadness}, \text{joy}, \text{disgust}\} \) emerging from a dataset using the NRC Word-Emotion Association Lexicon (EmoLex) \cite{bib:NCR}\footnote{While EmoLex provides a valuable resource for identifying emotion-related words, it has certain limitations. Specifically, it is based solely on word-level counts from the lexicon. It does not account for contextual factors such as negations, word dependencies, or the broader semantic structure of the text. Nevertheless, this approach remains meaningful, allowing for a systematic and consistent analysis of emotional expressions across texts, providing valuable insights into the overall emotional patterns within the dataset.}. Let \( P = \{p_1, p_2, \dots, p_N\} \) represent the set of posts in the dataset, where \( N \) is the total number of posts. For each post \( p_i \), we calculate the number of words associated with each emotion \( e \in E \), denoted by \( w_{e,p_i} \), where \( w_{e,p_i} \) is the number of words in post \( p_i \) that are associated with emotion \( e \). If a word is linked to multiple emotions, all associated emotions are considered. The proportion \( \rho_{e,p_i} \) of words in post \( p_i \) associated with emotion \( e \) is given by:
$\rho_{e,p_i} = \frac{w_{e,p_i}}{w_{p_i}},
$ where \( w_{p_i} \) is the total number of words in post \( p_i \) that are linked to any emotion.
At the dataset level, the average proportion of each emotion across all posts is computed as:
$
\bar{\rho}_e = \frac{1}{N} \sum_{i=1}^{N} \rho_{e,p_i}.
$

\item The indication of the presence of \textit{depression}. Let \( d_i \) represent the predicted depression label for post \( p_{i} \) (\( d_i \in \{ \text{no depression, moderate depression, severe depression} \} \)), as determined by the best-performing model from the Shared Task on \textit{Detecting Signs of Depression from Social Media Text} at LT-EDI-ACL2022~\cite{depression-detection}. We compute the proportion of each label \( l \in \{ \text{no depression, moderate depression, severe depression} \} \) at the dataset level as:
$
P(l) = \frac{N_{l}}{N},
$ where \( N \) is the total number of posts in the dataset and \( N_{l} \) the total number of posts in the dataset having the label $l$.

\end{itemize}
Additionally, given two datasets, we can identify the words that statistically distinguish them, using the methodology described in \cite{Monroe_Colaresi_Quinn_2017}, named the Fightin' Words. This approach is applied to find the words that differentiate the first dataset from the second and vice versa. Then we take the set of words that are statistically significant (having a \textit{z-score > 1.96}) in distinguishing a corpus with respect to another one. From the resulting list, we retain only the words associated with emotion in EmoLex to focus only on emotion-related terms.
Furthermore, to assess the semantic and thematic differences between these extracted words from both datasets, we tasked \gpt with categorizing them into distinct clusters. The model was given complete freedom to determine the labels and the number of clusters, with the only constraint being a maximum limit of 10 clusters. 
We chose to use a LLM for thematic clustering instead of traditional statistical or embedding-based methods because, unlike classical approaches that rely solely on co-occurrence patterns or fixed vector representations, LLMs draw on extensive contextual knowledge and semantic understanding gained through large-scale language training: this enables them to capture subtle relationships between words and to generate meaningful, human-interpretable labels for each identified cluster \cite{llm-clustering}. The obtained groups are thus directly interpretable.
The prompts used are in the Appendix \ref{appendix:clusters}. We then manually reviewed the resulting clusters by correcting any imprecise assignments. This approach allows us to analyze the extent to which the datasets differ regarding the themes they address.

Sentiment, emotion, depression analyses, z-scores, and thematic analysis provide quantitative insights into the emotional tone, mental health signals, and thematic patterns in self-descriptions, enabling comparisons with LLM-generated texts. Combined, these metrics help identify differences in portrayals and language use, offering a comprehensive view of how people with disabilities represent themselves versus how LLMs represent them.

\subsection{Addressing the RQs}


To answer RQ1, we apply the sentiment, emotion, and depression analysis on \llmdgem, \llmdgpt, \llmdmix, and \rdis and compare the values of the first three, i.e., the datasets generated by our LLMs, with those of people with disabilities (\rdis). Then, following \cite{Monroe_Colaresi_Quinn_2017}, we compare the words that distinguish \llmd from \rdis and vice versa.
To answer RQ2, we apply the sentiment, emotion, and depression analysis on the datasets generated by \gem
(\llmdgem and \llmndgem), \gpt (\llmdgpt and \llmndgpt), and \mix (\llmdmix and \llmndmix). We compute the differences between the datasets of each set. Finally, we compare the words that distinguish \llmd from \llmnd and vice versa, again using the methodology from \cite{Monroe_Colaresi_Quinn_2017}.
To summarize, Table \ref{tab:summary-models} reports the datasets we created and adopted to address the RQs.


\begin{table}[t]
\centering
\renewcommand{\arraystretch}{1.2}
\begin{tabular}{|l|c|c|c|c|c|}
\hline
Dataset & Author of the posts & Person & Person with Dis. & \# Posts & Avg. Tokens \\ 
\hline
\hline
\llmdgem  & \gem &  & \cmark & 360 & 243.01 \\ 
\llmndgem & \gem & \cmark &  & 360 & 179.63 \\ 
\hline
\llmdgpt  & \gpt &  & \cmark & 360 & 221.66 \\ 
\llmndgpt & \gpt & \cmark &  & 360 & 183.55 \\ 
\hline
\llmdmix  & \mix &  & \cmark & 360 & 247.97 \\ 
\llmndmix & \mix & \cmark &  & 360 & 180.90 \\ 
\hline
\hline
\rdis & Reddit users with dis. &  & \cmark & 1,250 & 207.55 \\ 
\llmd  & All LLMs &  & \cmark & 1,080 & 237.55 \\ 
\llmnd & All LLMs  & \cmark &  & 1,080 & 178.31 \\ 
\hline
\end{tabular}
\caption{Summary of datasets used in this study, indicating whether each set represents a generic person or a person with a disability, together with the number of posts and the average number of tokens per post. \textit{"All LLMs"} reported in the last two rows of the table refers to the dataset obtained by merging the posts generated by \gem, \gpt, and \mix.}
\label{tab:summary-models}
\end{table}

\section{Results and Discussions} \label{sec:results}

\subsection{RQ1: LLMs' portrayal of people with disability vs.\ self-descriptions on Reddit}\label{sec:results:rq1}

\begin{figure}[t]
    \centering
    \begin{subfigure}[b]{0.495\textwidth}
        \centering
        \includegraphics[width=\textwidth]{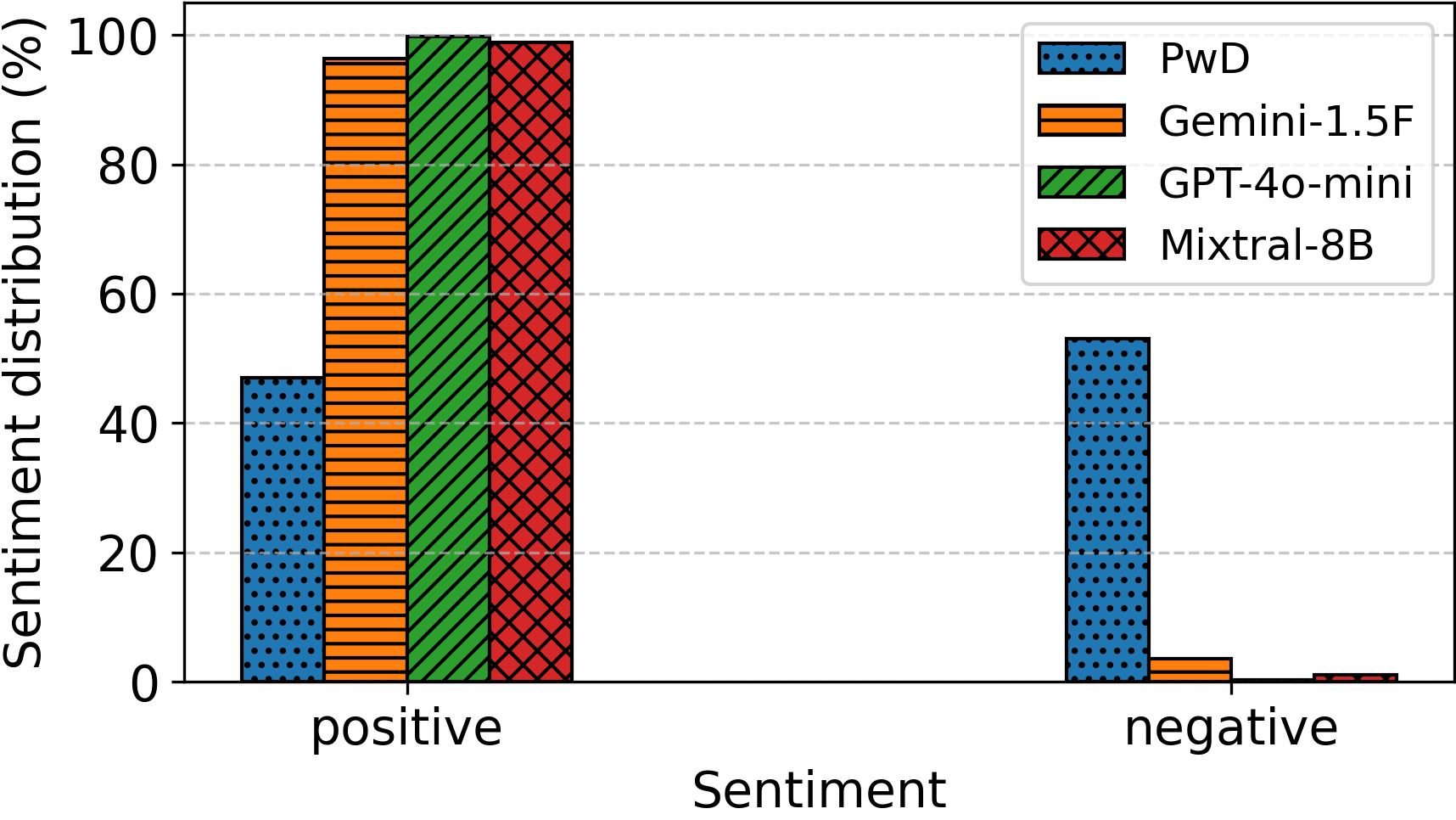}
        \caption{
        Sentiment analysis}
            \Description{Sentiment analysis on all the datasets. LLMs generated posts are almost all positive, while the majority of posts of RDIS are negative}
        \label{fig:sentiment-VADER}
    \end{subfigure}
    \hfill
    \begin{subfigure}[b]{0.495\textwidth}
        \centering
        \includegraphics[width=\textwidth]{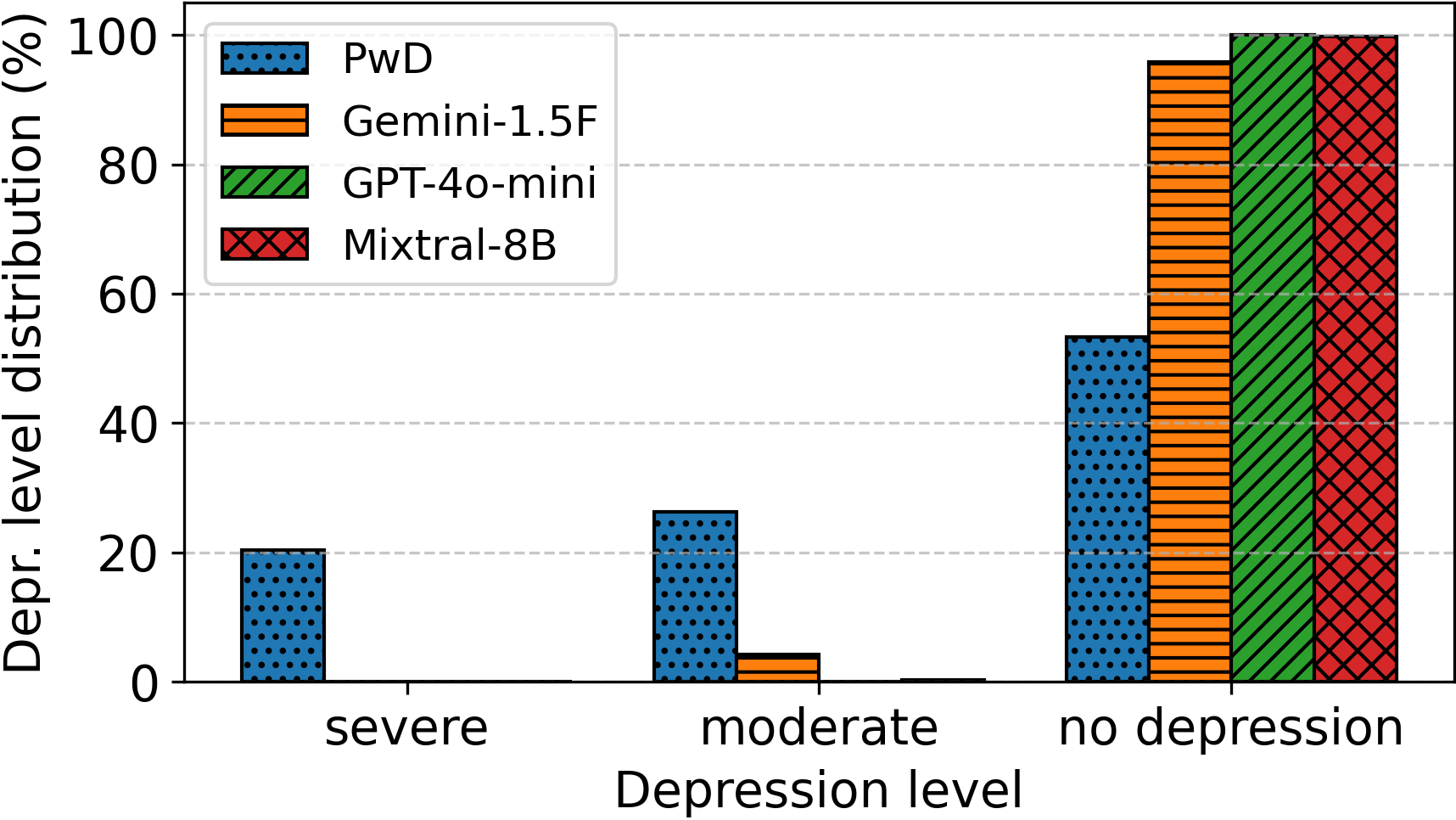}
        \caption{
        Depression analysis}
                    \Description{Depression analysis on all the datasets. LLMs-generated posts are almost all without depression, while the majority of posts of RDIS reveal severe or moderate depression.}
        \label{fig:depression-level}
    \end{subfigure}
    
    \vspace{1em}
    
    \begin{subfigure}[b]{0.85\textwidth}
        \centering
        \includegraphics[width=\textwidth]{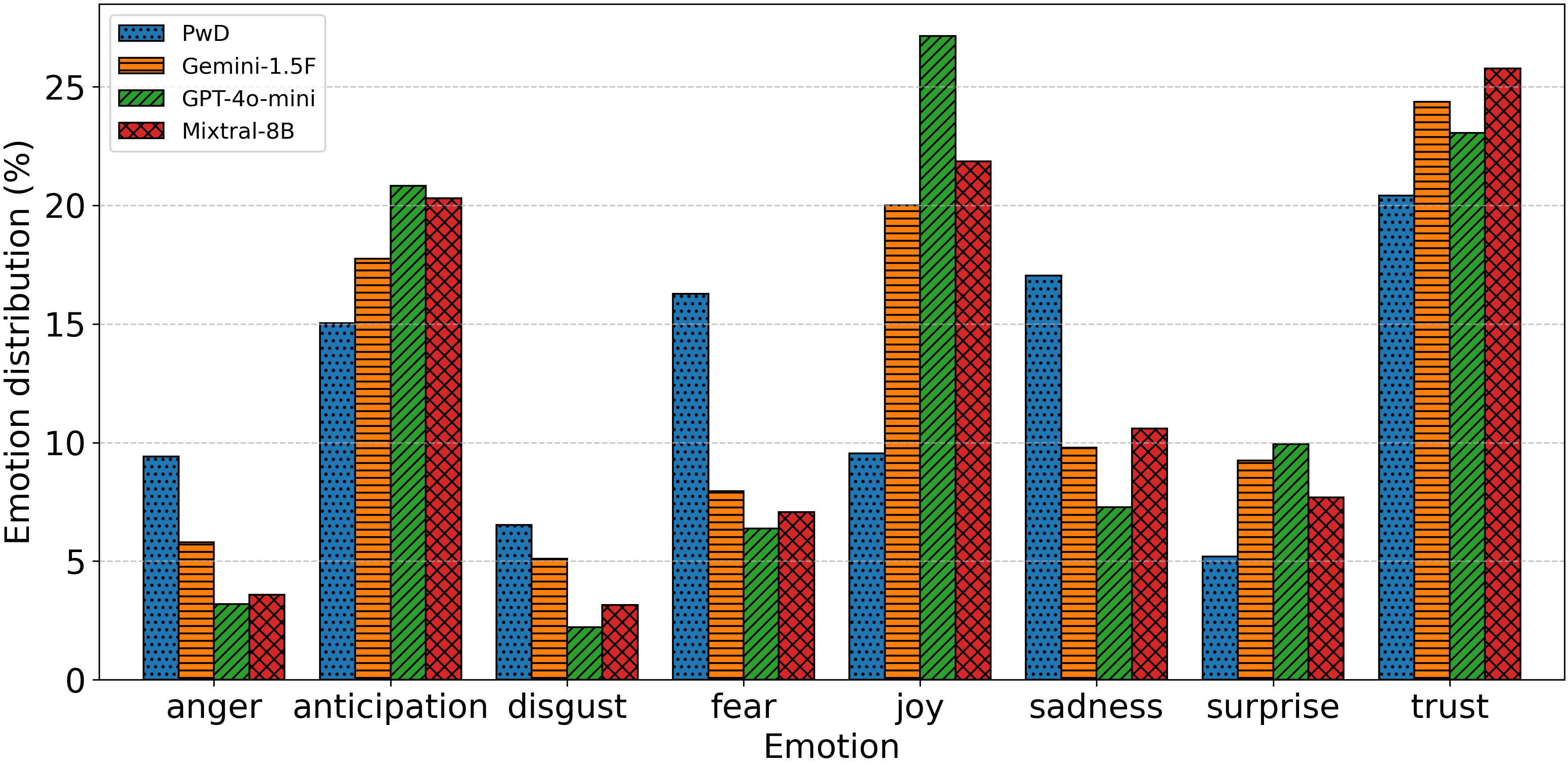}
        \caption{
        Emotion analysis.}
                    \Description{Emotion analysis on all the datasets. LLMs-generated posts have the highest percentage of positive emotions.}
        \label{fig:emotions-NCR}
    \end{subfigure}
    \caption{
    A comparison of sentiment, depression levels, and emotions between posts from the disability-related subreddits, i.e., written by people with disabilities (PwD, \rdis Dataset) and posts generated by \gem (\llmdgem Dataset), \gpt (\llmdgpt Dataset) and \mix (\llmdmix Dataset).
    }
    \label{fig:sentiment-depression-emotion}
\end{figure}

Figure \ref{fig:sentiment-depression-emotion} reports the differences between the posts in \rdis and each of the datasets of \llmd in terms of sentiment (Figure \ref{fig:sentiment-VADER}), depression level (Figure \ref{fig:depression-level}) and emotion (Figure \ref{fig:emotions-NCR}).

Referring to Figure \ref{fig:sentiment-VADER}, we see that the three LLMs predominantly suggest posts with a positive sentiment. The proportion ranges from $99.72\%$ for \gpt (\llmdgpt dataset) to $96.39\%$ for \gem (\llmdgem dataset). This contrasts starkly with the sentiment distribution of actual posts from Reddit users (\rdis dataset), where 
$53.06\%$
exhibits an overall negative sentiment. Similarly, Figure \ref{fig:depression-level} reveals that \gpt is devoid of any signs of depression, even when simulating the perspective of someone experiencing depression. Similarly, \mix presents only one post that exhibits moderate depression in \llmdmix dataset, while all the others are tagged with "not depression". For \gem, $4.17\%$ of its posts were identified as containing indications of "moderate depression", while $95.83\%$ of "not depression". These results contrast with the findings from posts from Reddit users where 
$20.42\%$
of posts show signs of "severe depression" and around 
$26.26\%$
exhibit signs of "moderate depression". Moreover, looking at emotions (Figure \ref{fig:emotions-NCR}), posts in \rdis, that is, authored by Reddit users, have a higher frequency of words related to negative emotions such as anger, disgust, fear and sadness. In contrast, posts generated by LLMs contain more words associated with positive emotions like joy, trust, surprise, and anticipation. These analyses reveal that LLMs tend to portray disability through an overly sweetened lens, emphasizing predominantly positive emotions. 

Table \ref{tab:top-words-llms-reddit} highlights the words that show a statistically significant difference (z-score > 1.96) between the \llmd and \rdis.  In contrast, Table \ref{tab:top-words-reddit-llms} presents the words that distinguish posts from \rdis compared to those of \llmd. Each word has been assigned a descriptive theme to understand the topics most considered in both posts from Reddit users and those LLM-generated.
Furthermore, to keep the tables compact, we report only the words linked to an emotion in EmoLex. However, the complete lists are reported in the Appendices \ref{full-topwords-llms-reddit} and \ref{full-topwords-reddit-llms}.

\begin{table}[t]
    \centering
    \begin{tabular}{|p{13.5cm}|}
       \hline
\textbf{Technology and Communication}: communication (6.52), technology (6.21), communicate (4.85) \\ \hline

\textbf{Disability and Advocacy}: \textit{inclusion (10.7)}, \textit{inclusive (9.31)}, blindness (8.15), accessible (6.93), cerebral (6.75), palsy (6.66), advocate (6.0), disability (4.71), differently (4.69), advocacy (4.13), stigma (3.14), inaccessible (2.61), helpful (2.6), supported (2.04), supporting (2.04) \\ \hline

\textbf{Community and Belonging}: \textit{share (19.06)}, \textit{community (17.06)}, \textit{love (10.51)}, embrace (6.49), join (4.99), kindness (4.89), familiar (4.29), diverse (3.95), acceptance (3.93), culture (3.57), found (3.52), contribute (3.28), trusty (3.17), joined (2.61), pride (2.38), isolated (2.35), building (2.17), cooperate (2.13), neighborhood (2.11), respectful (2.05), garden (1.97), network (1.96) \\ \hline

\textbf{Creativity and Expression}: \textit{unique (10.97)}, create (7.87), beauty (6.89), beautiful (6.57), art (6.24), creative (5.04), whirlwind (4.44), adventure (3.41), richness (3.12), hobby (2.98), fascinating (2.84), writer (2.79), music (2.69), engaging (2.66), crisp (2.41), shining (2.23) \\ \hline

\textbf{Motivation and Inspiration}:  \textit{forward (8.96)}, hope (7.9), passionate (7.24), important (7.06), importance (6.74), opportunity (5.82), encourage (5.39), inspire (4.44), continue (4.37), determination (4.27), finally (4.24), accomplishment (4.11), determined (4.0), focus (3.61), inspired (3.44), committed (3.41), passion (3.31), inspiration (3.23), uplift (3.22), ready (3.01), crucial (2.97), encouragement (2.67), eager (2.56), hopeful (2.32), granted (2.24), strive (2.23), tackle (2.23), effort (2.06) \\ \hline

\textbf{Joy and Enjoyment}:  \textit{excited (9.64)}, joy (8.78), favorite (7.42), wonderful (6.22), laughter (4.9), delicious (4.79), gratitude (4.69), sun (4.14), sunshine (3.7), enjoying (3.57), celebrating (3.08), refreshing (2.9), cherish (2.61), enjoy (2.44), fortunate (2.34), toast (2.34), aroma (2.28), lovely (2.24), exciting (2.23), laughing (1.98), celebrated (1.97), smile (1.97) \\ \hline

\textbf{Daily Life and Routine}:  small (8.05), spent (6.99), usual (4.99), routine (4.75), break (4.54), quiet (4.37), breakfast (4.11), smell (3.33), mountain (2.64), blanket (2.56), constant (2.44), staring (2.37), offer (2.31), daily (2.12), dinner (2.11) \\ \hline

\textbf{Learning and Growth}:  
\textit{journey (10.39)}, understanding (8.36), learn (7.12), learning (6.59), build (4.5), taught (3.89), progress (3.78), resources (3.65), grow (3.4), greater (2.9), growth (2.79), adapt (2.61), educate (2.54), essential (2.51), interesting (2.44), reading (2.28), explore (2.24), productive (2.14), plunge (2.13), interest (2.12), guide (1.98) \\ \hline

\textbf{Emotions and Mental States}:  \textit{grateful (9.34)}, proud (8.21), feeling (7.99), frustration (5.94), overwhelming (5.46), sense (4.49), overload (4.39), unexpected (3.58), surprisingly (3.27), valuable (3.03), comfort (2.91), soothing (2.9), peaceful (2.87), overwhelmed (2.85), depression (2.76), spirits (2.47), calm (2.22), appreciation (2.2), sadness (2.14), therapeutic (2.08), disheartening (2.05), exhaustion (2.04), empathy (2.03), apathy (1.97) \\ \hline

\textbf{Resilience and Strength}: tough (5.26), victory (4.59), strength (4.48), challenge (4.42), battle (3.4), powerful (3.38), resilient (3.01), patience (2.51), insurmountable (2.37)
\\
\hline
\end{tabular}
\caption{%
Statistically significant most distinctive words in \llmd compared to \rdis. We compute z scores following the method of \cite{Monroe_Colaresi_Quinn_2017} while retaining only words with emotions found in EmoLex and automatically clustering them across topics, which are sorted based on the average z-score of the terms. We highlight the top-10 words with the highest score in \textit{italic}.
}
\label{tab:top-words-llms-reddit}
\end{table}

\begin{table}[t]
    \centering
    \begin{tabular}{|p{13.5cm}|}
               \hline
\textbf{Education}: \textit{school (8.76)}, \textit{vision (6.62)}, university (4.33), fact (3.81), degree (3.39), teacher (2.76), study (2.7), level (2.44) \\ \hline

\textbf{Health and Illness}: \textit{pain (9.11)}, \textit{disabled (7.73)}, \textit{ill (6.18)}, sick (5.37), chronic (3.57), eat (3.15), food (2.64), painful (2.57), illness (2.52), dying (2.45), disease (2.28), cold (2.27), survive (2.25), clean (2.2), fat (1.98) \\ \hline

\textbf{Medical System and Experiences}: \textit{doctor (6.92)}, diagnosis (5.24), hearing (4.93), surgery (4.78), hospital (4.58), medical (4.34), aid (3.76), case (3.51), cancer (2.93), treat (2.59), nerve (2.58), specialist (2.39), accident (2.34), stroke (2.11), emergency (2.11), limp (2.03), mouth (1.98) \\ \hline

\textbf{Work and Career}: \textit{job (10.22)}, long (5.38), deal (4.71), start (4.46), quit (3.82), wear (3.67), leave (3.45), top (3.32), option (3.31), main (2.77), general (2.63), unemployed (2.52), plan (2.46), title (2.44), career (2.4), advance (2.4), major (2.28), management (2.25), solution (2.2), august (2.11) \\ \hline

\textbf{Mental Health and Emotions}: depressed (5.47), anxiety (4.81), thought (4.56), worried (3.99), disorder (3.78), upset (3.56), cry (3.54), crying (3.47), fear (3.44), angry (3.41), crazy (3.15), worry (3.06), stress (3.05), panic (2.89), mad (2.64), moderate (2.64), concerned (2.39), guilty (2.39), nervous (2.08), afraid (1.97) \\ \hline

\textbf{Negative States and Experiences}: \textit{worse (7.13)}, \textit{bad (7.1)}, \textit{hate (6.28)}, wrong (5.85), tired (5.78), wont (5.35), problem (5.29), shit (4.84), loss (4.42), hell (4.06), stupid (4.02), weird (3.94), hurt (3.87), fault (3.61), lose (3.48), losing (3.31), horrible (3.03), denied (2.98), fell (2.98), damage (2.96), lost (2.86), suffer (2.83), lower (2.77), blame (2.64), lazy (2.6), badly (2.58), suffering (2.52), miserable (2.52), terrible (2.46), failure (2.41), suddenly (2.38), useless (2.34), suck (2.25), ass (2.18), attack (2.18), hurting (2.18), awful (2.17), broken (2.16), damn (2.11), ruined (2.11), dumb (2.11), pathetic (2.06), annoying (2.06), falling (2.01), uncomfortable (2.0) \\ \hline

\textbf{Family and Relationships}: child (4.73), mother (4.68), brother (3.95), partner (3.73), mum (3.03), sex (2.87), father (2.85), god (2.52), childhood (2.36), baby (2.22), abuse (2.11) \\ \hline

\textbf{Financial and Legal Issues}: money (6.04), pay (4.81), afford (3.65), income (3.47), broke (3.28), legal (2.87), risk (2.76), government (2.58), limited (2.53), lawyer (2.45), debt (2.39), save (2.39), law (2.39), account (2.34), benefit (2.34), cover (2.24), homeless (2.18) \\ \hline

\textbf{Self-Harm and Suicide}: suicidal (3.52), kill (3.52), die (3.47), suicide (3.44), cutting (2.32), killing (2.32), death (2.03) \\ \hline

\textbf{Social and Communication}: advice (4.5), reason (4.44), happen (3.03), contact (2.83), vent (2.78), talk (2.75), coming (2.69), guess (2.69), fairly (2.64), letter (2.58), bother (2.57), speech (2.47), expect (2.28), obvious (2.2), actual (2.17), clue (2.11), fake (2.06), immediately (2.0), ignore (1.98) \\ \hline

    \hline
    \end{tabular}
\caption{Statistically significant most distinctive words in \rdis compared to \llmd. We compute z scores following the method of \cite{Monroe_Colaresi_Quinn_2017} while retaining only words with emotions found in EmoLex and automatically clustering them across topics, which are sorted based on the average z-score of the terms. We highlight the top-10 words with the highest score in \textit{italic}.}   
\label{tab:top-words-reddit-llms}
\end{table}

From the top distinctive words that statistically distinguish the LLM-generated posts from those of Reddit (Table \ref{tab:top-words-llms-reddit}) we note some very positive words like \textit{love}, \textit{grateful}, \textit{excited}, and \textit{joy}, \textit{proud}, \textit{passionate}, and \textit{gratitude}. These therefore predominantly appear in LLM-authored texts. These terms may reflect an idealized or exaggerated portrayal of emotional states and positivity, potentially indicative of LLMs aligning with socially desirable narratives rather than lived experiences. Furthermore, words such as \textit{share}, \textit{community}, and \textit{inclusion} suggest a strong emphasis on collective identity, advocacy, and social belonging. These terms are central to the discourse on disability advocacy and, therefore, appear to be emphasized by the models. Again, words such as \textit{inspire}, \textit{victory}, \textit{determination}, \textit{strength}, and \textit{battle} suggest a more inspirational and palatable narrative. The excessive use of these words has, however, been criticized in literature and news stories because it tends to create a one-dimensional portrayal of disability, only focusing on triumph over adversity rather than acknowledging the complex, multifaceted nature of lived experiences. This narrative has been called "inspiration porn" \cite{disability-centered_perspectives-2023} because reduces disability to a series of inspirational stories rather than acknowledging the real struggles, frustrations, and inequalities that disabled people face. 

Confirming this trend are the words that distinguish the posts written by Reddit users compared to those proposed by LLMs, listed in Table \ref{tab:top-words-reddit-llms}.  Terms like \textit{pain}, \textit{doctor}, \textit{hospital}, \textit{surgery}, and \textit{ill} are highly prevalent, reflecting the central role of medical and health-related issues in the lived experience of people with disability. The frequent appearance of these terms suggests that the disabled individuals’ discourse is deeply embedded in medical contexts. Words like \textit{job}, \textit{money}, \textit{unemployed}, \textit{pay}, and \textit{homeless} highlight significant economic and employment-related struggles for people with disabilities, issues underestimated by LLMs. Again, the frequent appearance of words such as \textit{bad}, \textit{worse}, \textit{hate}, \textit{stress}, \textit{anxiety}, \textit{suicidal} and \textit{crying} underlines the emotional and psychological burdens that come with disability. The use of words related to depression, anxiety, and emotional distress contrasts with the more sanitized emotional lexicon often found in LLM-generated texts. Words like \textit{suffering}, \textit{painful}, and \textit{hell} highlight the profound physical and emotional toll disability can take on individuals describing a visceral, raw, and unfiltered reality. Also, these issues seem to be excluded or romanticized in LLMs-generated texts.

\begin{figure}
    \centering
    \includegraphics[width=0.65\textwidth]{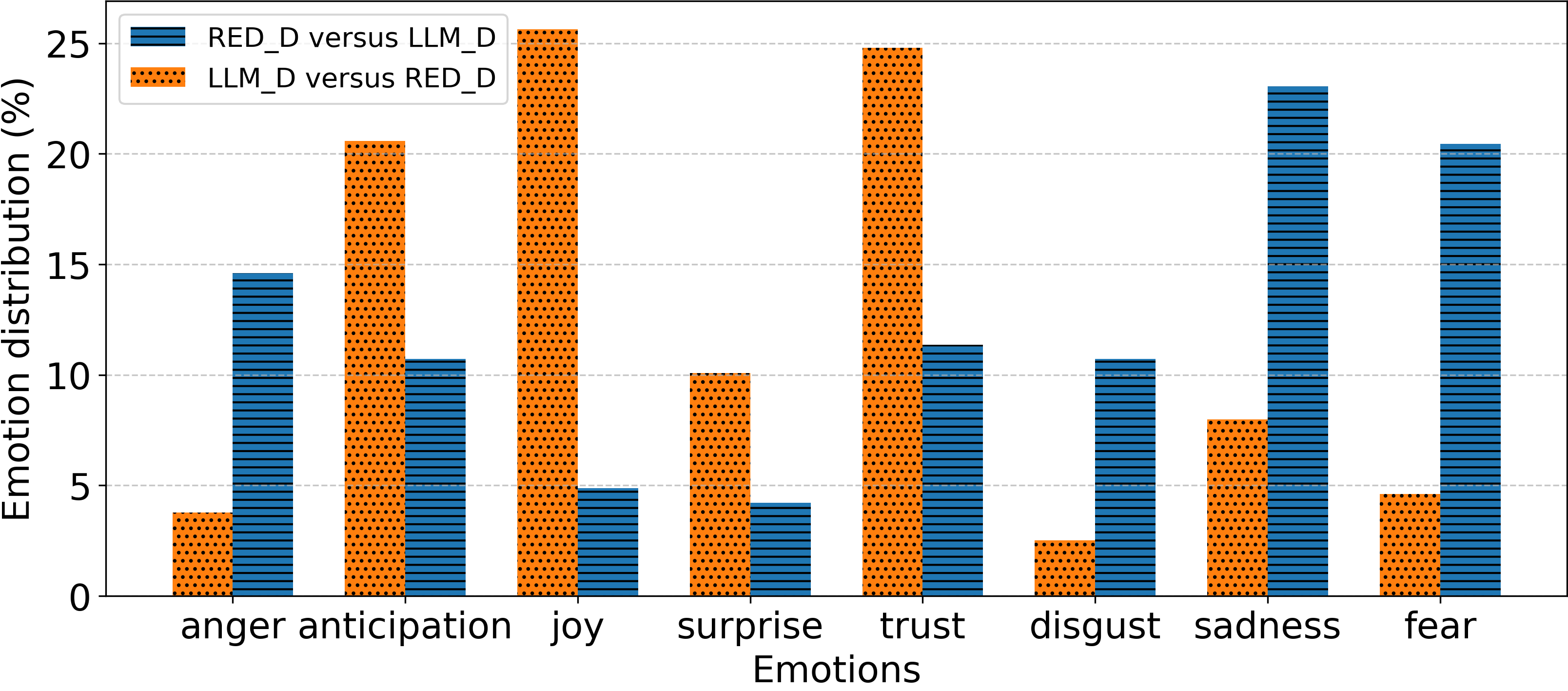}
    \caption{Emotional distributions of distinctive words in \llmd compared to \rdis and vice versa.}
        \Description{Plot showing the emotional distributions of distinctive words in RDIS compared to LLM_D and vice versa. LLM_D presents higher occurrences of negative emotions compared to RDIS.}
    \label{fig:emotional}
\end{figure}

Further confirmation of this analysis comes from the different categories assigned to each list of words. For Table \ref{tab:top-words-llms-reddit}, words have been clustered in 10 thematic groups related to themes of 
Technology and Communication, Disability and Advocacy, Community and Belonging, Creativity and Expression, Motivation and Inspiration, Joy and Enjoyment, Daily Life and Routine, Learning and Growth, Emotions and Mental States, Resilience and Strength.
For Table \ref{tab:top-words-reddit-llms}, words are categorized into 10 domains:  
Education, Health and Illness, Medical System and Experiences, Work and Career, Mental Health and Emotions, Negative States and Experiences, Family and Relationships, Financial and Legal Issues, Self-Harm and Suicide, Social and Communication.
The diverse categorization of distinctive words further highlights how disability is differently portrayed in texts written by Reddit individuals versus those generated by LLMs. While LLM-generated texts are dominated by aspirational and positive themes, focusing on emotions, growth, creativity, and overcoming obstacles, the language of individuals with disability is more grounded in the material, emotional, and legal challenges they face. The real experiences of disability —marked by medical struggles, socioeconomic hardships, legal battles, and emotional distress— are underrepresented in LLM-generated content, which tends to sanitize and idealize disability narratives.
Further evidence supporting this finding is presented in Figure \ref{fig:emotional}, which illustrates the proportional distribution of emotions associated with the words listed in Tables \ref{tab:top-words-llms-reddit} and \ref{tab:top-words-reddit-llms}. The distinctive words from LLM-generated content are disproportionately associated with positive emotions 
($81.09\%$),
such as joy, surprise, trust, and anticipation, compared to negative emotions 
($18.91\%$),
including anger, disgust, sadness, and fear. 
In contrast, the distinctive words in Reddit posts authored by disabled individuals are more strongly associated with negative emotions 
($68.82\%$)
than positive ones 
($31.18$\%).
Furthermore, an examination of individual emotions reveals that LLM-generated content exhibits higher proportions of all positive emotions than Reddit posts, whereas the opposite pattern emerges for negative emotions. This highlights a marked difference in the emotional tone of the two datasets.

\vspace{1em}
\noindent

\begin{table}[h!]
\begin{scriptsize}
\centering
\begin{tabular}{|p{13.5cm}|}
\hline
\textit{[...] I just feel like I've been in a constant state of being overwhelmed by everything. I'm so close to a meltdown at all times and I keep shutting down emotionally. I don't want to talk to anyone, go anywhere, or do anything. Having even a simple task that I need to do stresses me the hell out. Even thinking about it too much makes my skin crawl and I just want to shut down. I'm not really depressed, I just feel irritated by the entire world basically. [...]} 
\textbf{Post from \texttt{r/autism}}. \\ 
\hline
\textit{I was born blind. Always been this way. From the time I was in high school, I began to have really bad insecurities about my blindness. I've always gone to school with other blind people, but there were also sighted people there as well.. Growing up, I hated every blind person I went to school with. I couldn't relate to them at all, I couldn't find common ground, other than the fact we were all blind. By the time I got to high school, it just got worse and worse. I kept pointing the finger at how aweful these blind people were, when honestly, I was just pointing at myself. [...]}  
\textbf{Post from \texttt{r/blind}}. \\
\hline
\textit{[...] I'm crying every night and I've been contemplating suicide. I can't sleep and I barely eat. Along with that I struggle with ASD and ADHD. I can barely get out of bed in the morning. I've hung out with friends, gone on walks and so much more but I constantly feel the same way no matter what I do. I don't know how much longer I can continue. It feels like all the people that care about me are just doing it because they feel bad or it feels like they don't need me. [...]} \textbf{Post from \texttt{r/depression}}. \\
\hline
\hline
\textit{[...] For me, autism manifests in a few key ways.  Sensory sensitivities are a big one – loud noises can be incredibly overwhelming, and certain textures make me incredibly uncomfortable.  Bright lights are also a challenge.  I often need quiet time to recharge after social situations, even if I enjoyed them.  It's not that I *don't* like people, it's that social interactions require a lot of energy from me.
[...]
But I'm also incredibly proud of who I am.  My autism gives me unique perspectives and talents – I'm highly detail-oriented, I have a strong capacity for focus when I'm engaged in something I'm passionate about, and I see the world in a way many others don't.
I'm hoping to use this platform to share my experiences, advocate for autistic rights, and connect with other autistic individuals and allies. [...] Let's work together to create a more inclusive and understanding world. [...]} \textbf{Post generated by \gem when pretending to be a user with autism.} \\
\hline
\textit{[...] I’m a proud member of the blind community. I’ve been visually impaired since birth, and while my journey has had its challenges, it has also opened up incredible experiences and opportunities.
[...] One of my biggest passions is sharing my experiences and advocating for accessibility and inclusion. There’s so much work to be done to ensure that everyone, regardless of their abilities, has equal access to opportunities and spaces. It’s crucial that we create a world where everyone can thrive and feel valued.
I also want to highlight the amazing community I’ve found among fellow visually impaired individuals. We share stories, support one another, and inspire each other every day. It’s a reminder that connection is powerful, no matter the circumstances.
I hope to use this platform to educate, inspire, and share my journey. [...]} \textbf{Post generated by \gpt when pretending to be a blind user.} \\
\hline
\textit{[...] I know that talking about mental health can be difficult, but I believe that openness and honesty can help to reduce stigma and create a more supportive and inclusive community. For as long as I can remember, I have struggled with depression. It's a complex and multifaceted condition that has impacted many aspects of my life. Some days, it can be incredibly challenging to get out of bed, leave the house, or even engage in activities that once brought me joy. Other days, I may feel relatively ""normal"" and capable, only to be knocked down by a wave of sadness or hopelessness without warning.
[...] I have learned the importance of self-care, of reaching out for help when I need it, and of being patient and compassionate with myself as I work to manage my symptoms. I have also come to appreciate the incredible resilience and strength that exists within the human spirit.
Of course, there are still many days when I feel overwhelmed and unsure of how to move forward. But I am committed to continuing to prioritize my mental health, seeking out support and treatment as needed, and working to create a life that is meaningful and fulfilling despite the challenges I face.
If you or someone you know is struggling with depression, please know that you are not alone. [...]} \textbf{Post generated by \mix when pretending to be a user with depression.} \\
\hline
\end{tabular}
\caption{Example of posts' excerpt from people with a disability versus LLMs.}
\label{tab:examplerq1}
\end{scriptsize}
\end{table}

To further support this analysis, Table \ref{tab:examplerq1} contains examples of posts' excerpts written by real people versus those generated by the models. From the examples, it is clear that there are different sentiments, emotions, and tones between real people and LLMs. 
\modif{Therefore, the qualitative comparison also reveals a striking narrative discrepancy. Posts written by LLMs tend to conform to a standardized and uplifting narrative arc that can be described as a progression from \textit{challenge}, through \textit{growth}, to \textit{acceptance}. In this pattern, adversity is framed as a catalyst for personal development, culminating in a sense of meaning, resilience, or self-acceptance, as illustrated by the sentence \textit{"My autism gives me unique perspectives and talents"} generated by \gem and reported in Table \ref{tab:examplerq1}.
By contrast, many posts written by people with disabilities do not frame suffering as an opportunity for growth; instead, they often foreground exhaustion, frustration, and the ongoing difficulty of coping with disabling circumstances. Expressions of despair or endurance without resolution are common, emphasizing the chronic and unredeemed nature of their struggles, as in the sentence \textit{"I'm so close to a meltdown at all times and I keep shutting down emotionally"} from \texttt{r/autism} reported in Table \ref{tab:examplerq1}.}

\textbf{Answer to RQ1.} The results reveal that, in the context of a social media like Reddit, the LLMs' descriptions of disability significantly diverge from those from real people with disabilities. These descriptions are often idealized and overly optimistic, largely ignoring the negative emotions and challenges that individuals with disabilities face. Instead, they present a portrayal focused almost exclusively on positive sentiment and emotions, which risks fostering a form of toxic positivity that marginalizes those who are struggling and feeling overwhelmed by their disabilities, as warned by \cite{toxicpos2024}. 
What is worrying is the picture that emerges from the analysis of \rdis's posts: people with disabilities find themselves living very negative experiences that result in negative feelings and emotions due to the typical discomforts of disability sometimes expanded by a non-accessible and non-inclusive social-political system.

\subsection{RQ2: LLMs' portrayal of people with disability vs.\ generic individuals}

\begin{figure}
    \centering
    \begin{subfigure}{0.49\textwidth}
        \centering
        \includegraphics[width=\textwidth]{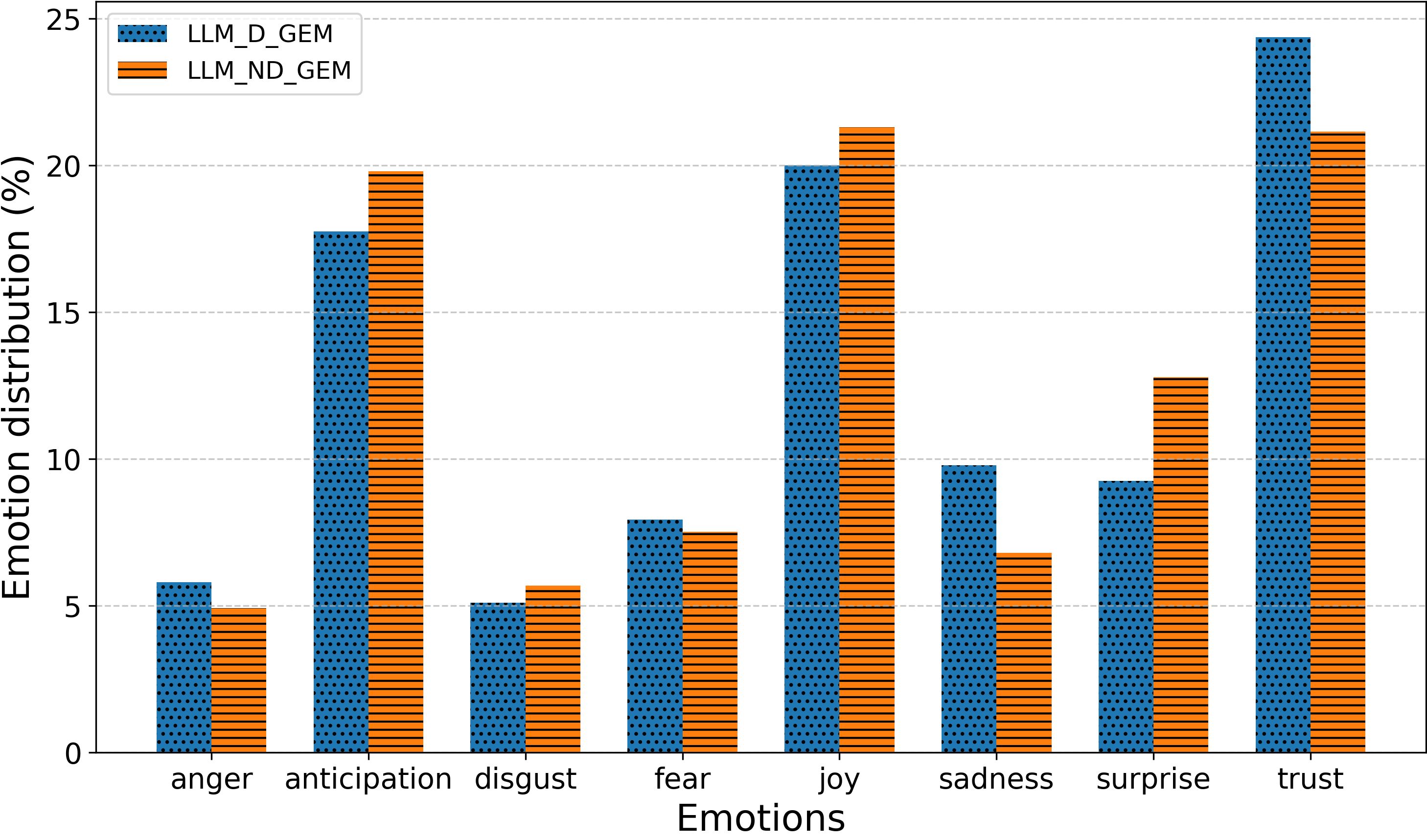}
        \caption{
        {\gem }
        }
            \Description{Emotion distribution in LLM_D_GEM and LLM_ND. For GEMINI-1.5F, LLM_ND present differences statistically significant (p-value < 0.05) for anger (-), anticipation (+), disgust (+), joy (+), sadness (-), surprise (+), and trust (-) with respect to LLM_D_GEM.}
        \label{fig:gemini-personas-NCR}
    \end{subfigure}
    \hfill
    \begin{subfigure}{0.49\textwidth}
        \centering
        \includegraphics[width=\textwidth]{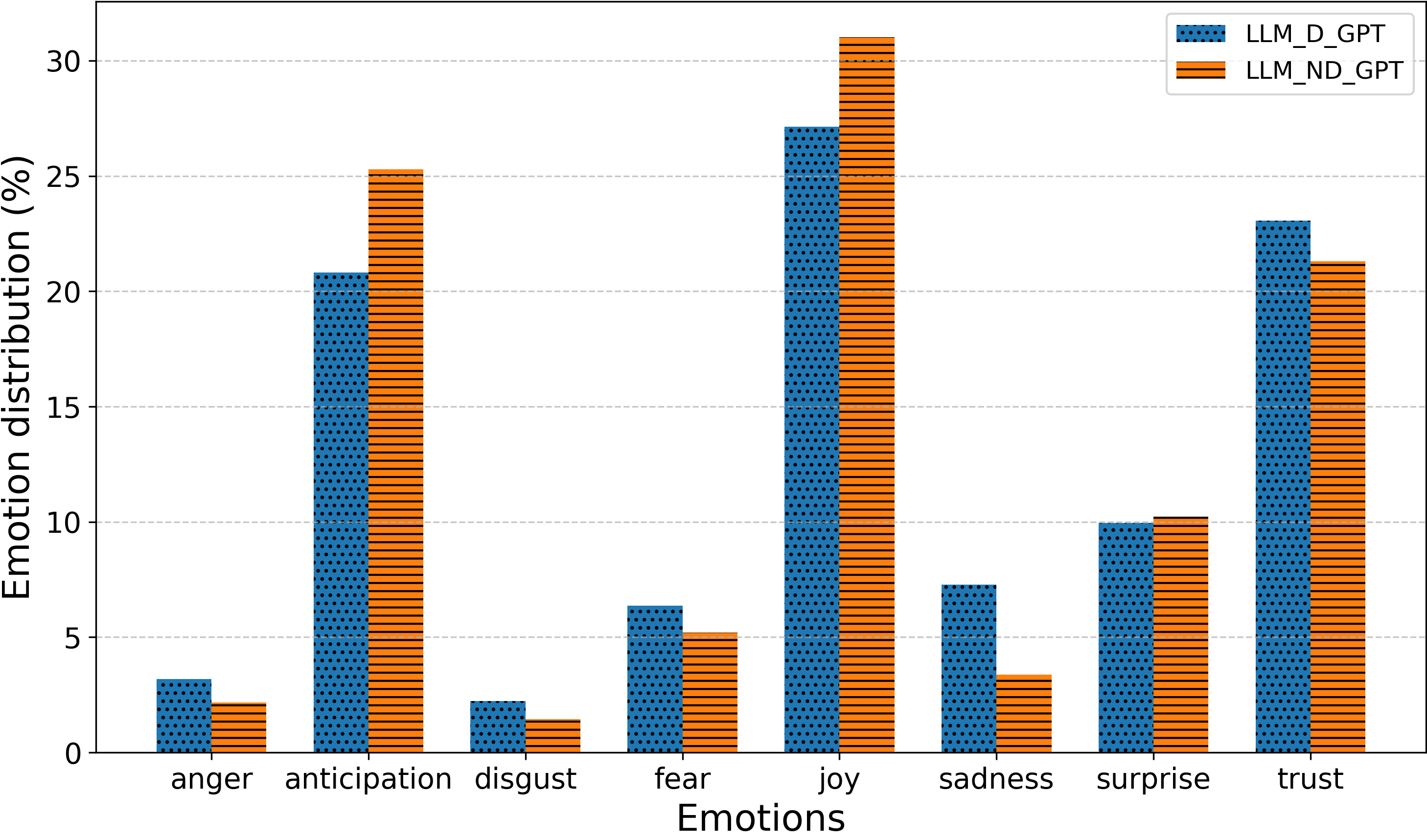}
        \caption{
        {\gpt}}
        \label{fig:gpt40mini-personas-NCR}
        \Description{Emotion distribution in LLM_D_GPT and LLM_ND. For GPT-4o-mini, LLM_ND present differences statistically significant (p-value < 0.05) anger (-), anticipation (+), disgust (-), fear (-), joy (+), sadness (-), and trust (-) with respect to LLM_D_GEM.}
    \end{subfigure}
    \vspace{1em}
    \begin{subfigure}{0.49\textwidth}
        \centering
        \includegraphics[width=\textwidth]{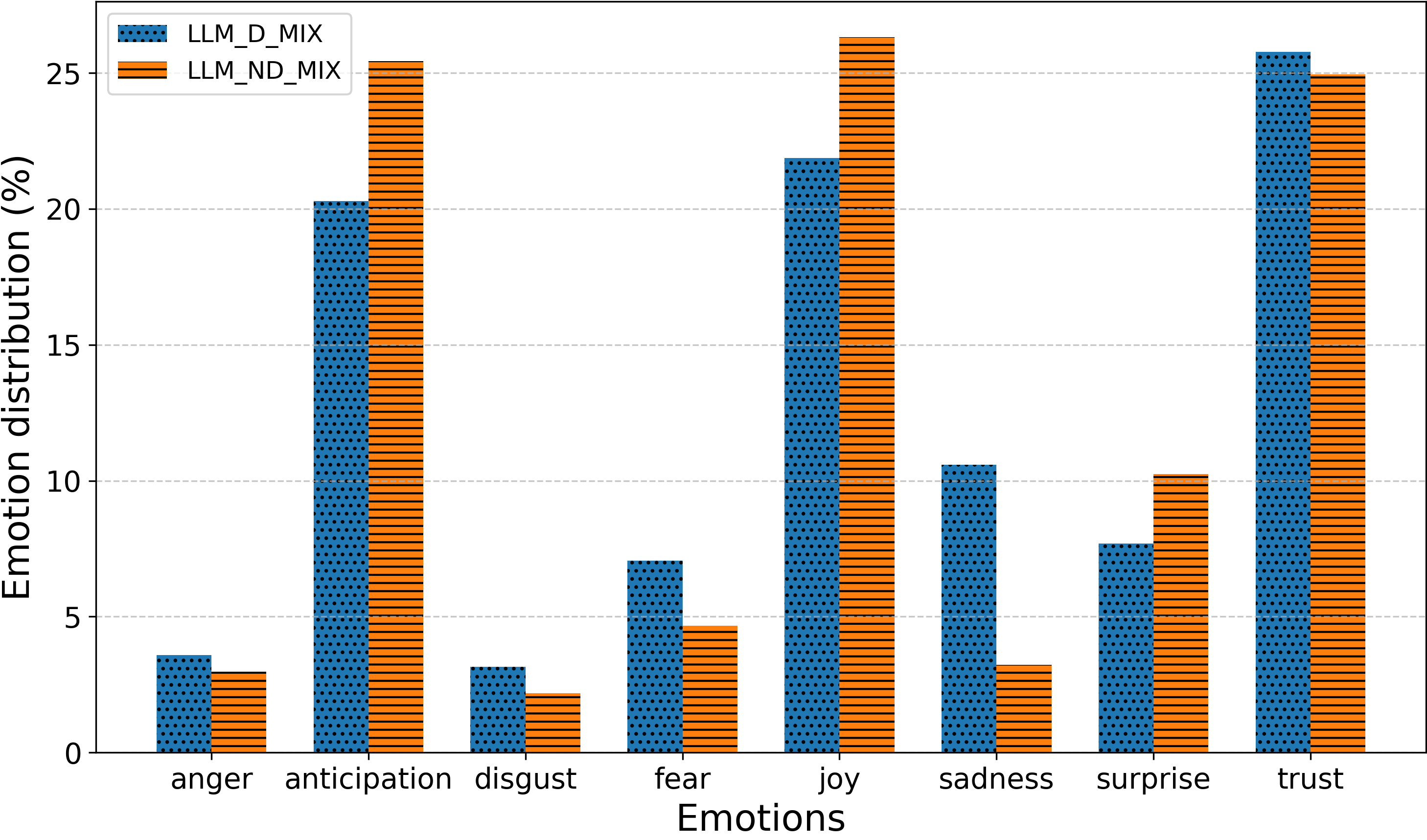}
        \caption{
        \mix}
        \label{fig:Mixtral-8B-personas-NCR}
        \Description{Emotion distribution in LLM_D_MIX and LLM_ND. For Mixtral-8B, LLM_ND present differences statistically significant (p-value < 0.05) for anticipation (+), disgust (-), fear (-), joy (+), sadness (-), and surprise (+) with respect to LLM_D_MIX.}
    \end{subfigure}
    \caption{
    Emotional distributions of posts generated by three different LLMs before and after specifying the presence of a disability. For \gem,  differences are statistically significant (p-value < 0.05) for  anger, anticipation, disgust,  joy, sadness, surprise, and trust. For \gpt, for anger, anticipation, disgust, fear, joy, sadness, and trust. For \mix, for anticipation, disgust, fear, joy, sadness, and surprise.
    }
    \Description{Emotional distribution of LLMs in LLM_D_GEM vs LLM_ND, LLM_D_GPT vs LLM_ND and LLM_D_MIX vs LLM_ND. The plots are described in each subfigure.}
    \label{fig:personasNCR}
\end{figure}

Figure \ref{fig:personasNCR} reports the emotional distributions of the words in posts generated by \gem (Figures \ref{fig:gemini-personas-NCR}), \gpt (Figure \ref{fig:gpt40mini-personas-NCR}) and \mix (Figure \ref{fig:Mixtral-8B-personas-NCR}) when prompted specifying whether the author of the post is a person with disability or not. The sentiment and depression plots are left to the Appendix (Figure \ref{fig:sentiment-persona-comparison} and \ref{fig:depression-persona-comparison} respectively) because the models propose almost exclusively positive and without depression posts in both settings.

Although the overall plots suggest a broadly similar emotional distribution across the models, distinct and consistent patterns emerge when examining the details.  For \gem, posts related to disability (\llmdgem) exhibit significantly higher levels of anger, lower anticipation, reduced disgust, less joy, greater sadness, decreased surprise, and increased trust compared to posts in \llmndgem.  \gpt reveals significant emotional differences between \llmdgpt and \llmndgpt, including higher anger, reduced anticipation, increased disgust, heightened fear, diminished joy, greater sadness, but enhanced trust. Likewise, \mix shows statistically significant distinctions between \llmdmix and \llmndmix, characterized by decreased anticipation, heightened disgust, greater fear, reduced joy, increased sadness, and diminished surprise. All these mentioned differences are statistically significant (p-value < 0.05) according to the t-test.
These findings indicate that while LLMs often idealize or oversimplify the experiences of disability, they still associate disability with more negative emotional tones compared to posts about individuals without disabilities. The only exceptions are reduced disgust (\gem) and increased trust (\gem and \gpt).

\begin{table}[t]
    \centering
    \begin{tabular}{|p{13.5cm}|}
       \hline
\textbf{Disability and Impairments}: \textit{disability (11.45)}, \textit{cerebral (9.22)}, \textit{palsy (9.21)}, \textit{blindness (7.68)}, cane (3.62), hearing (3.23), impairment (2.76) \\ \hline
\textbf{Inclusion and Accessibility}: \textit{inclusion (9.25)}, \textit{inclusive (7.87)}, \textit{accessible (6.72)}, public (3.59), extra (3.25), create (3.03), inaccessible (2.81) \\ \hline
\textbf{Strength and Resilience}: proud (5.94), struggle (4.72), tough (3.93), battle (3.49), strength (3.2), victory (3.03), determined (2.75), insurmountable (2.0) \\ \hline
\textbf{Acceptance and Understanding}: \textit{unique (7.8)}, \textit{understanding (6.93)}, differently (5.08), usual (4.7), acceptance (3.46), birth (3.01), culture (2.94), stigma (2.78), richness (2.52), greater (2.52), embrace (2.45), small (2.24), celebrating (2.06), deserve (2.05) \\ \hline
\textbf{Social Interactions and Communication}: communicate (5.6), communication (5.29), talk (3.88), share (3.59), familiar (3.18), explain (2.79), noise (2.17), staring (2.03) \\ \hline
\textbf{Emotions and Mental Health}: \textit{depression (9.19)}, weight (3.78), frustration (3.37), happy (2.7), loss (2.49), overwhelming (2.41), soothing (2.16), sadness (2.12), frustrated (2.07) \\ \hline
\textbf{Adaptation and Coping}: overload (4.09), routine (3.98), difficult (3.63), focus (3.15), break (2.97), constant (2.08), adapt (2.02) \\ \hline
\textbf{Support and Advocacy}: advocate (4.14), encourage (3.33), advocacy (3.28), helpful (3.2), guide (3.15), found (2.92), trusty (2.72), seek (2.27), granted (2.03) \\ \hline
\textbf{Isolation and Loneliness}: isolated (3.62), lack (3.06), isolation (2.11), darkness (2.01) \\ \hline
\textbf{Independence and Autonomy}: fully (3.09), manage (2.42), independence (2.32), full (2.2), build (2.13) \\ \hline
    \hline
    \end{tabular}
\caption{Statistically significant most distinctive words in \llmd compared to \llmnd. We compute z scores following the method of \cite{Monroe_Colaresi_Quinn_2017} while retaining only words with emotions found in EmoLex and automatically clustering them across topics, which are sorted based on the average z-score of the terms. We highlight the top-10 words with the highest score in \textit{italic}.}    \label{tab:top-words-personas-merged}
\end{table}

Table \ref{tab:top-words-personas-merged} highlights the words that show a statistically significant difference (z-score > 1.96) between the dataset of LLM-generated descriptions of disabled people vs.\ generic ones (\llmd and \llmnd). Conversely, Table \ref{tab:top-words-personas-merged_reverse} reports the words that significantly distinguish  \llmnd from \llmd datasets. Thematic clusters are also automatically identified using the same methodology from Section \ref{sub:comparison-metrics}. Table \ref{tab:top-words-personas-merged} contains terms related to inclusion, accessibility and advocacy, like \textit{inclusive} and \textit{advocate}, acceptance and understanding like \textit{embrace} and \textit{unique} and related to strength and resilience, like \textit{proud}, \textit{strength}, \textit{victory} and \textit{determined}. While these associations reflect positive traits and the agency of individuals with disabilities, it is again critical to contextualize such linguistic patterns to avoid perpetuating reductive narratives. Such portrayals can dehumanize individuals by defining them solely in relation to their challenges or their ability to "overcome" perceived limitations. In this context, terms like \textit{victory} and \textit{determined} must be critically examined to ensure they do not reinforce stereotypes that equate disability with suffering or frame people with disabilities as heroic for simply existing. 

\begin{table}[t]
    \centering
    \begin{tabular}{|p{13.5cm}|}
       \hline
\textbf{Entertainment and Fun}: \textit{pop (9.66)}, bookworm (5.41), memorable (3.39), fun (2.98), surprisingly (2.78), festival (2.32) \\ \hline
\textbf{Social and Relationships}: \textit{related (6.14)}, lurking (5.0), lover (4.55), enthusiast (4.26), fellow (2.55), kindness (2.05) \\ \hline
\textbf{Emotions and Feelings}: \textit{passionate (8.07)}, \textit{gratitude (7.45)}, \textit{excited (7.19)}, good (4.65), passion (4.6), feeling (3.96), content (3.87), excitement (3.54), perfect (3.39), chaos (3.23), forget (3.22), anxious (2.87), crazy (2.72), suddenly (2.71), eager (2.21), messy (2.21), sucker (2.13), heartfelt (2.13), hooked (2.04), luck (2.04), terrible (1.99) \\ \hline
\textbf{Creativity and Innovation}: inspiration (4.66), inspire (4.15), innovation (3.99), writer (2.9), captivating (2.0) \\ \hline
\textbf{Adventure and Exploration}: \textit{whirlwind (7.01)}, plunge (4.46), exciting (3.45), hidden (3.38), adventure (3.23), jump (3.2), catch (2.98), traveling (2.94), explore (2.94), wild (2.53), destination (2.22) \\ \hline
\textbf{Work and Career}: \textit{wait (7.22)}, \textit{status (5.7)}, \textit{relevant (5.69)}, productive (3.82), profession (3.74), productivity (3.39), spent (2.64), late (2.56), success (2.49), hit (2.48), machine (2.45), career (2.32), solution (2.12), designer (2.04), volunteer (2.03), tackle (2.01) \\ \hline
\textbf{Food and Dining}: delicious (4.28), food (3.19), dinner (2.88), crisp (2.6) \\ \hline
\textbf{Health and Wellbeing}: refreshing (4.03), sluggish (3.81), mindfulness (3.47), rest (2.25), chronic (2.02) \\ \hline
\textbf{Personal Growth and Development}: \textit{finally (5.68)}, forward (5.38), growth (4.81), personal (4.03), knowledge (3.55), thought (3.5), ahead (3.47), coming (3.06), inspired (2.89), completely (2.86), start (2.52), long (2.48), expect (2.35), real (2.34), accomplishment (2.29), intelligence (2.26), aspiring (2.24), grow (2.18), resist (2.13), learning (2.06), hobby (2.01) \\ \hline
\textbf{Nature and Beauty}: sunset (4.67), beautiful (4.14), bear (3.27), fall (2.84), dove (2.75), eagle (2.23), gorgeous (2.22), pretty (2.2), green (2.04), shelter (2.02), mountain (1.97) \\ \hline
    \hline
    \end{tabular}
\caption{Statistically significant most distinctive words in \llmnd compared to \llmd. We compute z scores following the method of \cite{Monroe_Colaresi_Quinn_2017} while retaining only words with emotions found in EmoLex and automatically clustering them across topics, which are sorted based on the average z-score of the terms. We highlight the top-10 words with the highest score in \textit{italic}.}   
\label{tab:top-words-personas-merged_reverse}
\end{table}

Instead, Table \ref{tab:top-words-personas-merged_reverse} shows several words related to (i) work and career, like \textit{profession}, \textit{career} and \textit{productivity}, and (ii) entertainment, adventure and pleasures of everyday life, such as \textit{sunset}, \textit{adventure}, \textit{explore}, \textit{festival} and \textit{dinner}. The contrastingly reduced usage of words related to these themes in the \llmnd dataset reflects a potential oversimplification of the lived experiences of individuals with disabilities. By prioritizing themes of resilience, advocacy, and accessibility, these texts may inadvertently frame disability as a monolithic identity defined by struggle, exclusion, or extraordinary effort while neglecting other fundamental aspects of human life, such as work and free-time activities. 

\begin{figure}
    \centering
    \includegraphics[width=0.75\textwidth]{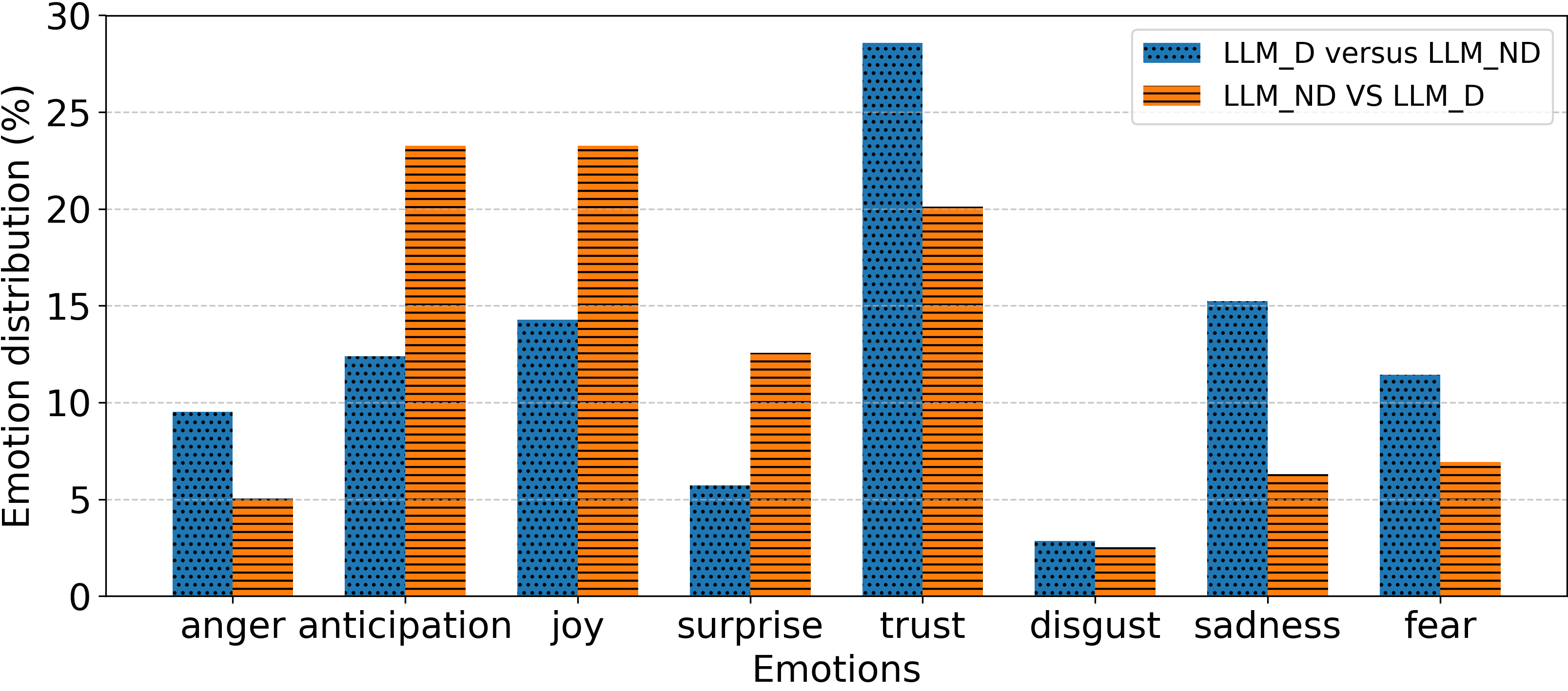}
    \caption{
    Emotional distributions of distinctive words in \llmd compared to \llmnd and vice versa.}
    \Description{Plot showing the emotional distributions of distinctive words in LLM_D compared to LLM_ND and vice versa. LLM_D presents higher occurrences of negative emotions compared to LLM_ND.}
    \label{fig:emotional_persona}
\end{figure}

Finally, Figure \ref{fig:emotional_persona} shows the distributions across EmoLex emotion of the words found in Tables \ref{tab:top-words-personas-merged} and \ref{tab:top-words-personas-merged_reverse}. In line with the previous results and discussion, the top words associated with people with disabilities present the highest proportion of negative emotions ($39.05\%$) than the top words associated with individuals without disabilities ($20.76\%$). Despite the models' tendency to idealize the portrayal of individuals with disabilities using words such as \textit{proud} and \textit{victory}, a closer analysis reveals that these highly positive words appear in contexts closely tied to words reflecting more negative emotions. A similar phenomenon was already observed in the context of gender and nationality \cite{marked-personas}.

Table \ref{tab:examplerq2} contains some examples of posts generated when LLMs pretend to be a general person. Manual analysis reveals that excerpts generated by the models, even when simulating posts from a "general person", often contain stereotypical portrayals and lack realism. While such overly positive depictions may seem benign in generic contexts, they risk being harmful when applied to sensitive topics, particularly in narratives involving people who are suffering, such as individuals with disabilities. In these cases, the tendency to simplify or idealize experiences can erase the complexities of real lives, contributing to misrepresentation and marginalization. Furthermore, qualitative analysis confirms the tendency of the model to talk about work and daily-life pleasure more when pretending to be people with disability.
\modif{Furthermore, from a narrative perspective, the \textit{challenge–growth–acceptance} arc observed in posts generated by LLMs simulating individuals with disabilities is largely absent from those simulating generic individuals. The latter tend to offer static, snapshot-like descriptions—often recounting the events of a single day—without the sense of progression or transformation that characterizes the disability-focused narratives. Examples of this type of posts are reported in Table \ref{tab:examplerq2}.}

\begin{table}[h!]
\begin{scriptsize}
\centering
\begin{tabular}{|p{13.5cm}|}
\hline
\textit{[...] Today was a pretty jam-packed day for me.  I started with a super energizing yoga session – definitely needed that to kickstart my morning! Afterwards, I tackled a mountain of laundry (the never-ending battle, am I right?! ). Then, I spent a few hours working on a freelance project –  fingers crossed it gets approved! This evening, I treated myself to a delicious homemade pasta dinner and caught up on my favourite podcast while enjoying the sunset.  Perfect ending to a perfect day. Looking forward to connecting with you all and sharing more about my life, adventures, and maybe even some recipe ideas! [...]} 
\textbf{Post generated by \gem when pretending to be a person.} \\ 
\hline
\textit{[...] Today was a pretty fantastic day! 
I started my morning with a strong cup of espresso and some journaling, which really set a positive tone for the day. After that, I decided to head out for a hike in the nearby nature reserve. The views were absolutely breathtaking, and I even spotted some adorable wildlife along the way! In the afternoon, I met up with a few friends for lunch at a new café in town. We had a blast catching up and trying their delicious avocado toast and homemade pastries  [...]}  
\textbf{Post generated by \gpt when pretending to be a person.} \\
\hline
\textit{I'm Marie, a nature lover and a lifelong learner. Today, I had an amazing day filled with fun and educational activities! I started my day by going on a refreshing hike in the nearby forest. The scenic views of the trees and wildlife were truly breathtaking, and it was a great way to connect with nature and get some exercise. Afterward, I visited a local museum to learn more about the history of my community. I was fascinated by the exhibits and gained a deeper appreciation for the people and events that shaped the place I call home.
Later in the day, I attended a virtual lecture on a topic that I've been eager to learn more about - sustainable living. The speaker shared practical tips and advice on how to reduce waste and conserve resources, and I'm excited to implement some of these strategies in my daily life.
Finally, I ended my day by cooking a delicious and healthy meal using locally sourced ingredients. It was a satisfying way to nourish my body and support local farmers at the same time. [...]} \textbf{Post generated by \mix when pretending to be a person.} \\
\hline
\hline
\end{tabular}
\caption{Example of posts' excerpt LLMs pretending to be a person.}
\label{tab:examplerq2}
\end{scriptsize}
\end{table}

\vspace{1em}
\noindent
\textbf{Answer to RQ2.} These results indicate that while LLMs tend to oversimplify and idealize disability experiences, they also associate disability with more negative emotional tones than posts about individuals without disabilities. Furthermore, the increased emphasis on words related to strength, resilience, and advocacy may unintentionally present disability as a reality defined by struggle, exclusion, and exceptional efforts to survive while overlooking other essential aspects of human life, such as career and entertainment. Portraying individuals solely through the lens of resilience and strength, and holding them up as sources of inspiration, risks perpetuating and normalizing the very conditions that demanded their resilience in the first place.

\section{Conclusions}
\label{sec:conclusions}
In this paper, we investigated how LLMs represent disability by comparing AI-generated portrayals with social media posts authored by individuals with disabilities. By leveraging a dataset of Reddit posts and corresponding artificial portrayals generated by LLMs, we analyzed the emotional tone, sentiment, and thematic coverage of these texts.

Our work contributes not only a publicly available dataset but also insights into the fundamental differences in how LLMs and real individuals describe disability, highlighting significant biases and oversimplifications in AI-generated representations. Most specifically, through our experiments we found that LLMs frequently idealize disability, producing overly positive and optimistic portrayals that ignore the complex realities and challenges faced by individuals with disabilities. In stark contrast, posts written by real individuals often convey more nuanced emotions, including negative feelings stemming from the intersection of their disabilities with inaccessible and non-inclusive societal systems. 
This disconnect underscores the risk of toxic positivity, where overly optimistic portrayals marginalize those struggling with the genuine hardships of living with a disability. 
\modif{These over-idealization likely stems from reinforcement learning from human feedback, safeguards and other alignment techniques that are explicitly designed to reduce outputs deemed offensive, harmful, or controversial. While well-intentioned, these guardrails can inadvertently encourage overly optimistic portrayals, as negative or ambivalent sentiments may be flagged as undesirable. Consequently, LLMs may learn to “err on the side of positivity”, producing content that appears respectful but glosses over the real challenges and complexities faced by individuals with disabilities.}

Moreover, we observed that LLMs adjust their language when simulating individuals with disabilities, associating disability-related posts not only with overly positive stereotypes but also with disproportionate negative emotional tones compared to posts about generic individuals. \modif{This is probably an implicit bias learned from training material collected from the Web where disability is often associated to negative emotions.}
The emphasis on terms like resilience and strength, while appearing commendable, \modif{if analyzed together with the more negative  emotional tones,} reduces individuals with disabilities to a narrative of struggle and exceptionalism. This framing risks perpetuating the exclusionary conditions that necessitate such resilience while failing to represent diverse aspects of their lives, such as professional achievements or recreational interests.

Our findings underscore the broader challenge of ensuring that LLMs represent marginalized groups authentically and respectfully. 
\modif{While efforts to address negative stereotypes in AI are critical, our study shows that positive overcorrections can also be harmful by oversimplifying and erasing the complexities of lived experiences related to disability.}
This calls for a more nuanced approach to mitigating bias in LLMs, one that acknowledges the diverse realities of marginalized groups without resorting to reductive idealizations.

In future work, we aim to \modif{further expand the qualitative analysis of the narrative structures present in generated and collected posts} and
extend our experiments to other languages and modalities to assess the prevalence of the behaviors we observed in LLMs and to encourage further research into the perception of disability through the lens of automatically generated text. We are particularly interested in exploring the limitations of group-based debiasing, as we believe that significant advancements in mitigating idealized portrayals in AI systems will require a more personalized, user-centric approach. 

\subsection*{Limitations}
\label{sec:limits}
Our analysis includes comparisons between Reddit content about people with disabilities and LLM-generated content, as well as between LLM-generated content about people with and without disabilities. 
\modif{Using publicly shared Reddit posts for research presents ethical risks: although the data are openly accessible, they contain sensitive personal experiences related to disability, mental health, or suffering, which could cause harm or distress if misinterpreted or decontextualized; furthermore, despite the anonymity of the posts, there remains a risk of inadvertently identifying individuals. However, this type of data is also used by reputable studies published in conferences and journals that are attentive to ethics and privacy \cite{ijcnlp2022Mondal} and we thus followed this practice.}

Moreover, as a potential limitation, we do not include a direct comparison between Reddit content about people with and without disabilities, which would parallel the LLM content analysis. This omission stems from the difficulty of reliably identifying Reddit content specifically about people without disabilities. Unlike discussions of disability—which are often explicit and concentrated in dedicated communities—references to non-disabled individuals are typically implicit and unmarked, making it challenging to retrieve representative or comparable data. We acknowledge this as a limitation and a potential direction for future research. Addressing it could further enrich our understanding of how LLMs represent disability relative to broader public discourse, without undermining the validity of the present findings.

Moreover, we have focused on a subset of disabilities to simplify the analysis: while this does not fully capture the complexity of the subject, it aligns with the approach taken in similar studies (c.f., \cite{gallegos24}, \cite{facct2024Glazko}). We then rely on lexicon-based tools to estimate emotions and sentiments and these tools may not fully account for contextual nuances; however, this methodology is also employed in authoritative studies to ensure the method remains explainable and reproducible \cite{marked-personas}. 

\modif{A further limitation of this study is the absence of human-in-the-loop evaluation to complement the quantitative analyses, particularly from individuals with disabilities. Although we include qualitative examples of the generated self-descriptions, these are not assessed by human annotators and therefore do not capture how such content may be interpreted or experienced by the communities represented. Incorporating people with disabilities as co-evaluators would provide more nuanced, context-sensitive insight and help ensure that the analysis reflects lived experience rather than solely computational measures. Future work should thus adopt participatory or community-informed evaluation approaches to more fully address the qualitative and ethical dimensions of LLM representations.}

Finally, these findings are specific to the versions of the models and the dates on which they were tested (especially those accessed via API). As LLMs are updated and their guardrails evolve, these results may change.


\subsection*{Ethical Considerations}
\label{sec:ethics}
Our research provides an in-depth exploration of how disability is represented in zero-shot LLMs. However, readers must consider the ethical implications of this topic. 
Existing debiasing and representation models only consider the “category” and not the “individual”, thus flattening and still stereotyping the representation and lacking the ability to tailor responses based on an individual's personal history, leading to potentially general, insensitive, or inappropriate responses. For example, one individual might appreciate being portrayed as a model of resilience and inspiration. In contrast, others might see it as disrespectful to the pain and discomfort one is forced to endure. The situation is complex and sometimes subjective, and a model that wants to be inclusive and respectful must know and conform to the experience of the person represented.
Similarly, these models seem incapable of adequately representing pain, suffering, and depression, often replacing them with positive language. While adopting an optimistic and hopeful tone may be appreciated in some contexts, in others, it can even prove harmful. Suffering is, in fact, an intrinsic part of human nature and accompanies each of us during various stages of life. Excluding it from these models would mean flattening and diminishing a fundamental aspect of the human experience. Rather than ignoring or masking it with positive words, it is necessary to acknowledge suffering and address it with empathy. This requires commitment both on a personal level, through mutual support, and on an institutional level, through inclusive and accessible policies. Reducing suffering must be a shared goal, but it cannot be achieved simply by avoiding the topic or replacing it with overly optimistic language.
Moreover, since these models are used on a large scale across numerous applications, it is clear that promoting an "only good vibes" approach risks marginalizing those who find themselves in situations of suffering. This risk does not only apply to people with disabilities but extends to anyone who is experiencing difficult moments in life.
Therefore, developing and using these models must consider the complexity of human emotions, representing them authentically and respectfully. Only in this way can we create tools capable of fostering genuine understanding, inclusion, and support. 
We acknowledge, however, that addressing such deeply personal and subjective topics may unintentionally fall short or cause misunderstandings. Our intention with this paper is not to offend or disrespect anyone, and we sincerely apologize if its content causes distress to any reader. Rather, our goal is to foster a constructive dialogue that bridges technologists and humanists, encouraging collaboration toward developing more inclusive AI systems.

\subsection*{Data Availability}
The code and the dataset used in our experiments are available at: \\
\url{https://github.com/marcobombieri/LLM-disability-representation}

\subsection*{Acknowledgments}
This research has received funding from: (i) the University of Mannheim's ``Gastwissenschaftler*-innenprogramm Nachhaltigkeit''; (ii) the MUR funded 2023-2027 Project of Excellence ``Inclusive Humanities: Perspectives for Development in the Research and Teaching of Foreign Languages and Literatures'' of the Department of Foreign Languages and Literatures of the University of Verona; (iii) the PNRR project FAIR - Future AI Research (PE00000013), under the NRRP MUR program funded by NextGenerationEU; (iv) the AI@TN2.0 project funded by the autonomous province of Trento.
Part of this work was carried out within the Digital Arena for Inclusive Humanities (DAIH) Research Centre at the University of Verona. The authors gratefully acknowledge this support.

\bibliographystyle{ACM-Reference-Format}
\bibliography{sample-base}

\newpage
\appendix
\section{Appendix}

\subsection{Prompt used for preprocessing the \rdis dataset} \label{appendix:prompt-used}
To extract from subreddits only posts written in the first person by people with disabilities describing personal experiences, we used the prompt reported below. Note that the choice of the optimal prompt is not crucial since this part was then refined manually as described in the paragraph \ref{sub:dataset-reddit}. This automatic preprocessing, therefore, has the sole purpose of simplifying the manual work needed to build the dataset, not replacing it. Follows the prompt used in this phase:

\begin{quote}
\textit{
    You are a text classifier operating on social media posts. You must classify posts into two disjoint classes, "1" or "2". Your answer must be in the format: "predictedClass;;explanation" where "predictedClass" can be "1" or "2" and "explanation" briefly describes why you have chosen that class. Separate "predictedClass" from "explanation" with the string ";;". Do not add other text. A post belongs to class "1" if: (the author of the post writes about himself/herself in the first person) AND ( the author of the post explicitly mentions his/her own disability/illness). A post belongs to class "2" otherwise. Follow the post you have to analyze:} 
    \\
    \{word\}
\end{quote}

\subsection{Prompts used for clustering a list of words} \label{appendix:clusters}
To assess the semantic and thematic differences between the extracted words with the method of \cite{Monroe_Colaresi_Quinn_2017}, we employed \gpt with the following prompt:

\begin{quote}
\textit{
    You will receive a list of words. Your task is to assign each word to a relevant topic. Do not propose more than 10 topics. Please ignore the numbers in parentheses in the list.} 
    \\
    \{words\}
\end{quote}

Then, once obtained the relevant topics, we used \gpt with the following prompt to classify each words in each cluster:
\begin{quote}
\textit{
        I have a list of words, and I would like you to classify them into the following topics: \\
        \{topics\} \\        
        Please carefully examine the meaning of each word and assign it to the most appropriate topic. 
        If a word could fit into multiple categories, choose the one where it is most relevant.
        In the answer, do not add extra text or clarification; only report the topic.
        Classify this word: \\
        \{word\}
}
\end{quote}
\gpt is used with temperature set to $1.0$ in the first prompt and $0.0$ in the second.

\subsection{Table \ref{tab:top-words-llms-reddit} not filtered with EmoLex}
\label{full-topwords-llms-reddit}
\begin{spacing}{0.8}
{\footnotesize 
share (19.06), a (17.7), everyone (17.67), world (17.25), community (17.06), experiences (15.15), today (14.7), alex (14.24), name (14.22), challenges (14.06), bit (14.02), moment (13.62), introduce (13.35), connect (13.35), accessibility (13.19), day (12.74), and (12.12), lets (12.07), wanted (11.77), sharing (11.64), others (11.58), your (11.07), little (11.06), unique (10.97), hey (10.79), inclusion (10.7), love (10.51), journey (10.39), our (10.25), with (10.2), about (10.1), hello (10.05), its (9.86), excited (9.64), amazing (9.6), support (9.58), take (9.58), grateful (9.34), inclusive (9.31), moments (9.27), the (9.26), navigating (9.25), you (9.1), of (9.07), youre (9.01),  (8.97), forward (8.96), navigate (8.95), joy (8.78), learned (8.63), each (8.48), understanding (8.36), ways (8.31), proud (8.21), felt (8.19), connecting (8.17), blindness (8.15), small (8.05), feeling (7.99), together (7.95), hope (7.9), create (7.87), awareness (7.8), morning (7.73), names (7.65), cerebralpalsy (7.53), connection (7.44), favorite (7.42), part (7.39), resilience (7.31), myself (7.29), challenging (7.29), passionate (7.24), incredible (7.22), who (7.2), afternoon (7.18), stories (7.12), learn (7.12), important (7.06), perspective (7.05), spent (6.99), accessible (6.93), beauty (6.89), reminded (6.77), free (6.75), cerebral (6.75), importance (6.74), some (6.72), thoughts (6.71), reach (6.69), palsy (6.66), learning (6.59), barriers (6.57), beautiful (6.57), communication (6.52), embrace (6.49), those (6.42), we (6.35), emotions (6.26), woke (6.24), art (6.24), wonderful (6.22), technology (6.21), remind (6.17), advocate (6.0), todays (5.97), vibrant (5.97), filled (5.97), sometimes (5.96), frustration (5.94), celebrate (5.93), coffee (5.9), deaf (5.84), victories (5.83), opportunity (5.82), inclusivity (5.8), faced (5.78), perspectives (5.69), okay (5.69), experience (5.68), chat (5.68), simple (5.68), always (5.68), disabilityawareness (5.66), language (5.65), advocating (5.63), difference (5.6), taking (5.59), differences (5.54), park (5.51), obstacles (5.49), overwhelming (5.46), deafcommunity (5.46), sincerely (5.45), define (5.43), encourage (5.39), deafness (5.38), sensory (5.36), despite (5.36), believe (5.36), disabilitypride (5.31), to (5.3), tough (5.26), thrilled (5.25), autismawareness (5.24), living (5.24), autism (5.23), thank (5.21), often (5.18), individuals (5.13), mix (5.12), inclusionmatters (5.09), coordination (5.08), fresh (5.08), looking (5.07), whether (5.06), reminder (5.05), air (5.05), creative (5.04), sign (5.03), space (5.03), positivevibes (5.02), join (4.99), usual (4.99), positive (4.96), face (4.95), neurodiversity (4.91), way (4.91), laughter (4.9), kindness (4.89), later (4.88), frustrating (4.86), communicate (4.85), fulfilling (4.85), movement (4.83), abilities (4.82), nature (4.8), delicious (4.79), eg (4.79), leaves (4.77), easy (4.77), affects (4.77), rollercoaster (4.76), routine (4.75), many (4.75), spaces (4.74), hear (4.73), finds (4.73), disability (4.71), tasks (4.71), incredibly (4.69), can (4.69), birds (4.69), gratitude (4.69), differently (4.69), supportive (4.65), adaptive (4.62), assistive (4.62), opportunities (4.59), victory (4.59), selfcare (4.59), express (4.58), deep (4.56), break (4.54), new (4.54), build (4.5), sense (4.49), strength (4.48), diversity (4.45), whirlwind (4.44), inspire (4.44), virtual (4.43), challenge (4.42), managed (4.39), overload (4.39), continue (4.37), quiet (4.37), also (4.33), through (4.32), conversation (4.31), familiar (4.29), alone (4.29), determination (4.27), youarenotalone (4.26), finally (4.24), shaped (4.22), dedefness (4.22), creativity (4.2), post (4.16), downs (4.16), strengths (4.15), calming (4.15), sun (4.14), all (4.14), feels (4.14), senses (4.13), advocacy (4.13), appreciate (4.11), breakfast (4.11), accomplishment (4.11), colors (4.1), communitysupport (4.1), cues (4.08), particularly (4.07), triumphs (4.06), shared (4.06), suggestions (4.06), conversations (4.04), determined (4.0), days (4.0), isolating (3.98), mentalhealth (3.98), diverse (3.95), finding (3.95), deafawareness (3.94), painting (3.94), regardless (3.94), acceptance (3.93), member (3.93), fullest (3.91), joining (3.91), taught (3.89), evening (3.88), inspiring (3.87), promote (3.86), overcome (3.85), helps (3.85), power (3.85), creating (3.84), media (3.83), passions (3.82), warmth (3.78), exploring (3.78), progress (3.78), events (3.77), accessibilitymatters (3.77), listening (3.76), means (3.75), brings (3.74), mention (3.73), equitable (3.73), great (3.72), sunshine (3.7), misconceptions (3.7), yoga (3.69), event (3.69), interests (3.69), myday (3.69), fantastic (3.68), heavy (3.66), cup (3.66), tea (3.65), resources (3.65), person (3.65), different (3.63), catching (3.63), struggling (3.61), focus (3.61), ups (3.61), find (3.6), monumental (3.6), signlanguage (3.6), unexpected (3.58), rewarding (3.58), enjoyed (3.58), enjoying (3.57), culture (3.57), designed (3.56), tools (3.56), strategies (3.56), mixed (3.56), sending (3.54), raise (3.53), activities (3.53), online (3.52), found (3.52), glimpse (3.49), stereotypes (3.46), inspired (3.44), gentle (3.43), committed (3.41), adventure (3.41), open (3.4), grow (3.4), battle (3.4), powerful (3.38), joys (3.38), struggles (3.38), book (3.36), society (3.36), wave (3.33), smell (3.33), positivity (3.33), encountered (3.33), smallvictories (3.32), sensoryoverload (3.32), lift (3.31), passion (3.31), contribute (3.28), mentalhealthawareness (3.27), chirping (3.27), surprisingly (3.27), helped (3.25), meaningful (3.24), inspiration (3.23), may (3.23), uplift (3.22), allows (3.21), attended (3.21), wind (3.21), local (3.18), disabilities (3.18), representation (3.18), thrive (3.18), platform (3.18), connections (3.17), focused (3.17), highlight (3.17), embracing (3.17), trusty (3.17), grey (3.17), connected (3.14), stigma (3.14), raising (3.13), project (3.13), bag (3.12), richness (3.12), perseverance (3.12), breeze (3.12), pottery (3.12), textures (3.1), tiny (3.09), celebrating (3.08), fan (3.08), wheelchairuser (3.06), ramp (3.06), dailylife (3.06), eventful (3.06), solutions (3.06), read (3.05), welcoming (3.05), websites (3.05), brighter (3.05), valuable (3.03), sounds (3.03), fog (3.03), more (3.02), hi (3.02), ready (3.01), wheelchairlife (3.01), mentalhealthmatters (3.01), satisfying (3.01), journeys (3.01), resilient (3.01), participate (3.0), relaxing (3.0), reaching (2.99), hobby (2.98), coding (2.97), crucial (2.97), sarah (2.96), baking (2.96), visuallyimpaired (2.96), flowers (2.96), hesitate (2.94), headed (2.94), tips (2.93), session (2.93), news (2.93), reflecting (2.92), valued (2.92), comfort (2.91), remember (2.91), reflect (2.9), tricky (2.9), greater (2.9), lego (2.9), refreshing (2.9), deafandproud (2.9), differentnotless (2.9), soothing (2.9), simplest (2.89), rustling (2.89), interactions (2.88), rights (2.88), peaceful (2.87), surrounded (2.85), overwhelmed (2.85), fascinating (2.84), slice (2.84), caf (2.84), fluorescent (2.84), autismacceptance (2.84), blindandproud (2.84), wins (2.84), biggest (2.84), rich (2.84), podcast (2.83), yeah (2.82), spectrum (2.81), bright (2.8), uplifting (2.79), everydaylife (2.79), writer (2.79), nevergiveup (2.79), growth (2.79), communitylove (2.79), meeting (2.77), depression (2.76), come (2.75), lipreading (2.75), huge (2.74), step (2.73), itsokaytonotbeokay (2.73), trees (2.73), american (2.73), cooking (2.72), foster (2.72), grocery (2.72), for (2.71), situations (2.71), music (2.69), story (2.67), chatting (2.67), blindnessawareness (2.67), beige (2.67), continuing (2.67), encouragement (2.67), growing (2.66), id (2.66), engaging (2.66), own (2.66), battles (2.64), mountain (2.64), social (2.63), asl (2.63), belonging (2.63), insert (2.63), adapt (2.61), inaccessible (2.61), joined (2.61), audiobook (2.61), hum (2.61), nutshell (2.61), cherish (2.61), deaflife (2.61), luna (2.61), blindcommunity (2.61), lights (2.6), helpful (2.6), heres (2.59), started (2.58), eager (2.56), blanket (2.56), crowded (2.56), spread (2.56), climbing (2.55), rustle (2.54), liam (2.54), educate (2.54), overall (2.54), recharge (2.53), adventures (2.53), essential (2.51), shapes (2.51), patience (2.51), value (2.51), hoping (2.51), visual (2.51), strong (2.5), matters (2.5), engage (2.5), win (2.49), communities (2.49), needs (2.49), heavier (2.48), unwind (2.48), adaptability (2.48), cozy (2.48), tapestry (2.48), thats (2.47), spirits (2.47), puzzle (2.45), constant (2.44), interesting (2.44), enjoy (2.44), bridge (2.44), encounter (2.44), comforting (2.43), dressed (2.43), listened (2.43), disabilitycommunity (2.41), navigated (2.41), crisp (2.41), smells (2.41), questions (2.38), pride (2.38), surrounding (2.38), staring (2.37), environments (2.37), insurmountable (2.37), facing (2.37), isolated (2.35), bed (2.35), homemade (2.35), oatmeal (2.35), counts (2.35), scent (2.35), goto (2.35), mystory (2.35), overlook (2.35), fortunate (2.34), toast (2.34), might (2.34), yearold (2.33), hopeful (2.32), offer (2.31), movie (2.31), latest (2.31), predictable (2.3), things (2.29), youve (2.29), reading (2.28), aroma (2.28), satisfaction (2.28), deafpride (2.28), insights (2.25), slightly (2.25), explore (2.24), granted (2.24), lovely (2.24), ourselves (2.24), shop (2.24), strive (2.23), exciting (2.23), shining (2.23), tackle (2.23), assumptions (2.23), calm (2.22), autumn (2.2), hurdles (2.2), prioritize (2.2), empowerment (2.2), appreciation (2.2), spreading (2.2), misunderstandings (2.2), sights (2.2), fantasy (2.18), wrap (2.18), similar (2.18), pure (2.18), yours (2.18), organizations (2.17), building (2.17), highlights (2.16), ensure (2.16), empowered (2.16), wash (2.16), tackled (2.16), truly (2.15), typical (2.14), productive (2.14), pushing (2.14), sadness (2.14), capable (2.13), managing (2.13), deeply (2.13), plunge (2.13), imagination (2.13), unspoken (2.13), sheer (2.13), squirrels (2.13), artists (2.13), cooperate (2.13), laughs (2.13), focusing (2.13), daily (2.12), interest (2.12), meaning (2.11), workout (2.11), written (2.11), sound (2.11), dinner (2.11), primarily (2.11), neighborhood (2.11), hobbies (2.1), touch (2.1), us (2.1), definitely (2.09), finished (2.08), therapeutic (2.08), promoting (2.08), keen (2.08), shifted (2.08), while (2.07), effort (2.06), singing (2.05), disheartening (2.05), signing (2.05), diving (2.05), worlds (2.05), caught (2.05), deafculture (2.05), muted (2.05), recipe (2.05), puzzles (2.05), playlist (2.05), clouds (2.05), respectful (2.05), selfadvocacy (2.05), freshly (2.05), overlooked (2.05), winding (2.05), nuances (2.05), lows (2.05), supported (2.04), exhausting (2.04), supporting (2.04), exhaustion (2.04), empathy (2.03), certainly (2.03), that (2.03), ideas (2.02), complex (2.02), freelance (2.0), predictability (2.0), defined (2.0), specific (2.0), laughing (1.98), towards (1.98), guide (1.98), hopefully (1.98), favourite (1.98), volunteering (1.98), made (1.97), guidedog (1.97), smoothie (1.97), bustling (1.97), chamomile (1.97), garden (1.97), spark (1.97), swirling (1.97), nonetheless (1.97), noisecancelling (1.97), celebrated (1.97), attitudes (1.97), apathy (1.97), smile (1.97), neurological (1.97), supermarket (1.96), network (1.96), projects (1.96), distant (1.96), meditation (1.96), colleagues (1.96), followed (1.96)
}
\end{spacing}

\subsection{Table \ref{tab:top-words-reddit-llms} not filtered with EmoLex}
\label{full-topwords-reddit-llms}
\begin{spacing}{0.8}
{\footnotesize{

i (20.4) do (15.23) dont (14.98) because (14.1) get (12.59) they (12.53) she (12.43) he (11.92) cant (11.64) have (11.44) years (11.34) now (11.08) no (10.92) very (10.56) so (10.55) her (10.38) only (10.37) job (10.22) go (10.14) would (9.59) since (9.37) at (9.36) year (9.11) pain (9.11) just (9.02) when (8.88) told (8.86) school (8.76) back (8.63) him (8.58) said (8.48) if (8.35) will (8.34) months (8.25) last (8.07) it (8.02) never (8.01) didnt (7.93) them (7.83) disabled (7.73) ago (7.72) think (7.6) need (7.51) point (7.47) should (7.46) not (7.41) over (7.4) work (7.38) got (7.36) an (7.28) what (7.23) anymore (7.2) parents (7.15) worse (7.13) like (7.12) but (7.11) bad (7.1) two (7.07) could (7.03) doctor (6.92) college (6.89) his (6.76) first (6.67) old (6.64) want (6.63) vision (6.62) anyone (6.55) diagnosed (6.55) did (6.52) having (6.31) hate (6.28) going (6.26) left (6.19) ill (6.18) past (6.17) issues (6.14) until (6.13) normal (6.1) asked (6.07) before (6.06) as (6.05) in (6.04) money (6.04) why (6.02) anything (6.0) ive (5.99) does (5.87) wrong (5.85) really (5.83) again (5.82) right (5.82) tired (5.78) tell (5.74) eye (5.73) mild (5.64) know (5.63) ear (5.56) mom (5.53) recently (5.52) enough (5.52) stuff (5.51) scared (5.49) depressed (5.47) used (5.47) nothing (5.44) due (5.4) high (5.4) havent (5.39) long (5.38) sick (5.37) doctors (5.36) wont (5.35) problem (5.29) fucking (5.28) try (5.26) diagnosis (5.24) sure (5.23) was (5.21) put (5.15) almost (5.14) month (5.13) week (5.12) sorry (5.1) adhd (5.1) still (5.09) severe (5.07) end (5.06) basically (4.95) jobs (4.94) after (4.93) hearing (4.93) had (4.93) self (4.92) ever (4.91) next (4.91) problems (4.91) already (4.9) home (4.87) car (4.87) weeks (4.85) extremely (4.84) shit (4.84) house (4.84) is (4.81) pay (4.81) anxiety (4.81) wondering (4.8) any (4.79) surgery (4.78) idea (4.77) child (4.73) able (4.72) deal (4.71) doing (4.7) saying (4.69) drive (4.69) mother (4.68) situation (4.66) lot (4.65) seems (4.64) how (4.63) dad (4.58) hospital (4.58) thought (4.56) gotten (4.52) guys (4.52) either (4.5) advice (4.5) away (4.48) me (4.47) barely (4.46) start (4.46) reason (4.44) aids (4.43) give (4.42) loss (4.42) fuck (4.4) says (4.4) medical (4.34) being (4.33) university (4.33) f (4.29) second (4.28) state (4.17) girl (4.14) off (4.11) physically (4.09) etc (4.09) stop (4.09) half (4.08) three (4.08) adult (4.07) insurance (4.06) hell (4.06) sleep (4.04) relationship (4.03) stupid (4.02) kid (4.02) older (4.02) called (4.0) worried (3.99) guy (3.96) sort (3.96) brother (3.95) apply (3.94) weird (3.94) hurts (3.9) starting (3.9) hurt (3.87) probably (3.87) legally (3.86) m (3.86) issue (3.83) quit (3.82) became (3.82) fact (3.81) obviously (3.8) disorder (3.78) side (3.78) man (3.78) aid (3.76) partner (3.73) out (3.73) move (3.73) symptoms (3.73) cause (3.72) leg (3.69) wear (3.67) student (3.67) gone (3.67) afford (3.65) much (3.63) done (3.63) figure (3.63) mentally (3.63) anyway (3.62) legs (3.61) fault (3.61) getting (3.6) though (3.6) chronic (3.57) this (3.57) upset (3.56) meds (3.56) into (3.55) cry (3.54) spastic (3.52) kids (3.52) suicidal (3.52) kill (3.52) ok (3.52) case (3.51) lose (3.48) context (3.47) crying (3.47) die (3.47) income (3.47) people (3.46) yet (3.46) gonna (3.45) leave (3.45) fear (3.44) suicide (3.44) nobody (3.44) head (3.43) younger (3.43) ssdi (3.43) idk (3.43) been (3.41) angry (3.41) sister (3.41) classes (3.41) degree (3.39) entire (3.39) therapist (3.39) once (3.39) fine (3.39) too (3.38) test (3.38) lol (3.38) bipolar (3.38) yesterday (3.38) nearly (3.36) husband (3.36) literally (3.35) couldnt (3.33) covid (3.33) top (3.32) option (3.31) losing (3.31) cannot (3.31) close (3.3) sit (3.3) wasnt (3.3) broke (3.28) edit (3.28) term (3.28) s (3.28) taken (3.27) gets (3.26) asking (3.26) absolutely (3.25) without (3.24) office (3.24) wish (3.24) girlfriend (3.24) handle (3.23) kinda (3.22) waiting (3.22) supposed (3.21) straight (3.2) my (3.2) currently (3.19) wanna (3.19) reddit (3.19) sucks (3.19) age (3.18) came (3.18) wants (3.18) type (3.17) hoh (3.17) talking (3.16) theyre (3.15) brain (3.15) crazy (3.15) eat (3.15) hes (3.15) bc (3.14) ssi (3.14) applied (3.14) wouldnt (3.1) on (3.1) reasons (3.1) unless (3.1) middle (3.08) fucked (3.08) wife (3.08) four (3.08) worry (3.06) moved (3.06) walking (3.05) stress (3.05) whenever (3.05) worst (3.04) phone (3.04) mum (3.03) dating (3.03) horrible (3.03) write (3.03) happen (3.03) family (3.03) most (3.01) few (3.01) explained (2.99) turn (2.98) denied (2.98) hated (2.98) exam (2.98) foot (2.98) fell (2.98) damage (2.96) fast (2.96) between (2.95) run (2.95) especially (2.95) approved (2.93) ptsd (2.93) boss (2.93) causing (2.93) audiologist (2.93) cancer (2.93) poor (2.93) actually (2.91) program (2.91) stuck (2.9) hour (2.89) needed (2.89) panic (2.89) th (2.89) multiple (2.89) front (2.89) everything (2.88) terrified (2.87) sex (2.87) thinks (2.87) paid (2.87) country (2.87) legal (2.87) anyways (2.87) rent (2.87) applying (2.87) hard (2.87) lost (2.86) same (2.86) business (2.85) father (2.85) speak (2.85) shes (2.85) trouble (2.84) stopped (2.84) during (2.83) contact (2.83) benefits (2.83) sitting (2.83) normally (2.83) suffer (2.83) couple (2.82) boyfriend (2.82) grade (2.82) likely (2.8) shouldnt (2.8) wake (2.78) vent (2.78) caused (2.78) english (2.77) lower (2.77) main (2.77) shut (2.77) knew (2.77) female (2.76) risk (2.76) teacher (2.76) uk (2.76) sub (2.76) dropped (2.76) effects (2.76) buy (2.76) except (2.76) gave (2.76) possible (2.76) then (2.75) talk (2.75) pass (2.74) service (2.73) usually (2.72) amount (2.72) somewhere (2.72) else (2.71) paperwork (2.7) possibly (2.7) surgeries (2.7) confused (2.7) socially (2.7) study (2.7) trauma (2.7) male (2.7) far (2.7) happened (2.7) coming (2.69) twice (2.69) oh (2.69) guess (2.69) under (2.69) appointments (2.68) hours (2.67) arm (2.66) considering (2.66) known (2.65) fairly (2.64) mad (2.64) moderate (2.64) adults (2.64) subreddit (2.64) crutches (2.64) blame (2.64) married (2.64) extreme (2.64) none (2.64) tldr (2.64) graduated (2.64) qualify (2.64) food (2.64) training (2.63) awkward (2.63) becoming (2.63) begin (2.63) led (2.63) failed (2.63) general (2.63) treated (2.63) call (2.61) research (2.6) soon (2.6) someone (2.6) driving (2.6) recommended (2.6) bunch (2.6) causes (2.6) lazy (2.6) housing (2.6) well (2.59) treat (2.59) whatever (2.58) care (2.58) room (2.58) nerve (2.58) testing (2.58) christmas (2.58) period (2.58) government (2.58) math (2.58) tests (2.58) badly (2.58) letter (2.58) syndrome (2.58) tried (2.58) painful (2.57) telling (2.57) receive (2.57) bother (2.57) summer (2.57) conditions (2.57) unfortunately (2.57) random (2.54) somewhat (2.54) repeat (2.54) yall (2.54) limited (2.53) tend (2.52) god (2.52) nor (2.52) ex (2.52) suffering (2.52) higher (2.52) amp (2.52) campus (2.52) bullshit (2.52) miserable (2.52) abusive (2.52) unemployed (2.52) anybody (2.52) info (2.52) essentially (2.52) where (2.52) illness (2.52) happens (2.51) date (2.5) forever (2.5) feel (2.5) grew (2.47) severely (2.47) speech (2.47) several (2.46) plan (2.46) treatment (2.46) terrible (2.46) dying (2.45) medications (2.45) senior (2.45) private (2.45) ocd (2.45) grades (2.45) lawyer (2.45) license (2.45) medication (2.45) control (2.44) level (2.44) anywhere (2.44) title (2.44) eyes (2.43) realized (2.42) woman (2.42) passed (2.42) security (2.41) failure (2.41) results (2.41) must (2.41) career (2.4) advance (2.4) fix (2.4) eventually (2.4) debt (2.39) coworkers (2.39) dead (2.39) alot (2.39) save (2.39) burnout (2.39) specialist (2.39) position (2.39) drivers (2.39) law (2.39) concerned (2.39) guilty (2.39) hemiplegia (2.39) affected (2.38) suddenly (2.38) appointment (2.37) bathroom (2.36) childhood (2.36) sad (2.35) account (2.34) least (2.34) accident (2.34) retina (2.34) asd (2.34) knee (2.34) useless (2.34) cried (2.34) benefit (2.34) expensive (2.34) cutting (2.32) killing (2.32) drug (2.32) mistakes (2.32) semester (2.32) diplegia (2.32) graduate (2.32) flare (2.32) third (2.31) mentioned (2.31) paying (2.31) instead (2.3) willing (2.29) apartment (2.29) looked (2.28) major (2.28) expect (2.28) disease (2.28) feet (2.28) regular (2.27) tells (2.27) bills (2.27) wrote (2.27) directly (2.27) cold (2.27) leaving (2.27) thinking (2.26) cost (2.25) states (2.25) hardly (2.25) pills (2.25) ankle (2.25) reasonable (2.25) hr (2.25) financial (2.25) officially (2.25) attacks (2.25) everybody (2.25) management (2.25) yo (2.25) ci (2.25) suck (2.25) survive (2.25) trans (2.25) yelling (2.25) ignored (2.25) stand (2.25) cover (2.24) mainly (2.24) students (2.22) drink (2.22) floor (2.22) baby (2.22) cut (2.21) background (2.21) wearing (2.21) obvious (2.2) solution (2.2) psychiatrist (2.2) calling (2.2) clean (2.2) functioning (2.2) sleeping (2.2) or (2.2) accept (2.19) retail (2.18) manager (2.18) prior (2.18) genetic (2.18) x (2.18) department (2.18) joint (2.18) siblings (2.18) six (2.18) ass (2.18) october (2.18) pt (2.18) attack (2.18) bone (2.18) damaged (2.18) pissed (2.18) women (2.18) nowhere (2.18) besides (2.18) switch (2.18) died (2.18) hurting (2.18) clinic (2.18) homeless (2.18) rude (2.17) blood (2.17) tested (2.17) stressed (2.17) awful (2.17) actual (2.17) say (2.17) single (2.16) broken (2.16) shoes (2.16) burden (2.14) posted (2.13) correct (2.13) use (2.12) damn (2.11) clue (2.11) ruined (2.11) dumb (2.11) neurologist (2.11) abuse (2.11) coworker (2.11) himself (2.11) lay (2.11) application (2.11) august (2.11) pump (2.11) classmates (2.11) stroke (2.11) mg (2.11) wedding (2.11) google (2.11) emergency (2.11) faking (2.11) stayed (2.11) non (2.1) arms (2.1) rant (2.1) personally (2.1) raised (2.1) nervous (2.08) wonder (2.08) line (2.08) answers (2.08) dream (2.08) theyd (2.06) pathetic (2.06) round (2.06) annoying (2.06) glasses (2.06) response (2.06) fake (2.06) allowed (2.05) short (2.05) young (2.05) changed (2.04) kept (2.03) ends (2.03) pm (2.03) wheelchairs (2.03) january (2.03) stage (2.03) limp (2.03) embarrassed (2.03) yrs (2.03) death (2.03) medicare (2.03) studies (2.03) ppl (2.03) fired (2.03) turns (2.03) abused (2.03) migraines (2.03) zero (2.03) becomes (2.03) undiagnosed (2.03) pair (2.03) drugs (2.03) hand (2.03) function (2.03) considered (2.03) affecting (2.03) volume (2.03) apparently (2.02) above (2.02) personality (2.01) weak (2.01) sensitive (2.01) falling (2.01) based (2.01) processing (2.01) aware (2.01) beginning (2.0) immediately (2.0) uncomfortable (2.0) cp (2.0) up (2.0) relate (1.99) fat (1.98) nerves (1.98) theyll (1.98) uni (1.98) meltdowns (1.98) hip (1.98) ignore (1.98) gait (1.98) studying (1.98) aspergers (1.98) mouth (1.98) afraid (1.97) fit (1.97) large (1.96) 
}}
\end{spacing}

\subsection{Table \ref{tab:top-words-personas-merged} not filtered with EmoLex}
\label{full-topwords-personas-merged}
\begin{spacing}{0.8}
{\footnotesize
i (14.44), accessibility (11.51), disability (11.45), my (11.38), not (10.6), deaf (10.36), autism (9.48), inclusion (9.25), cerebral (9.22), palsy (9.21), depression (9.19), am (8.77), have (8.76), others (8.72), who (8.59), person (7.89), that (7.88), inclusive (7.87), unique (7.8), blindness (7.68), sometimes (7.63), alone (7.55), is (7.53), challenges (7.47), be (7.16), understanding (6.93), like (6.82), but (6.74), accessible (6.72), navigate (6.64), many (6.6), can (6.58), if (6.54), cerebralpalsy (6.52), autistic (6.29), disabilities (6.21), world (6.14), awareness (6.1), okay (6.1), want (6.05), help (6.03), its (6.0), sign (5.98), proud (5.94), blind (5.8), sensory (5.77), affects (5.72), struggling (5.62), doesnt (5.62), might (5.62), communicate (5.6), thank (5.59), barriers (5.54), youre (5.53), feel (5.51), often (5.47), deafness (5.44), communication (5.29), for (5.27), names (5.24), may (5.18), are (5.11), learned (5.09), live (5.08), differently (5.08), cp (5.08), because (5.02), inclusivity (4.98), movement (4.95), wheelchair (4.92), understand (4.9), asl (4.86), disabilityawareness (4.86), which (4.86), coordination (4.84), sincerely (4.8), advocating (4.75), take (4.72), struggle (4.72), ways (4.72), language (4.7), hard (4.7), usual (4.7), feels (4.69), deafcommunity (4.69), differences (4.67), more (4.67), also (4.66), days (4.59), me (4.56), bed (4.54), means (4.54), sounds (4.53), despite (4.52), frustrating (4.51), spaces (4.51), autismawareness (4.5), define (4.48), even (4.48), disabilitypride (4.46), ask (4.43), abilities (4.38), inclusionmatters (4.37), part (4.31), heavy (4.3), neurodiversity (4.27), mobility (4.27), than (4.26), experiences (4.25), assistive (4.25), visual (4.19), advocate (4.14), adaptive (4.1), overload (4.09), there (4.07), condition (4.0), own (3.99), routine (3.98), please (3.97), helps (3.97), needs (3.94), tough (3.93), using (3.91), reach (3.9), navigating (3.9), physical (3.88), talk (3.88), celebrate (3.88), obstacles (3.87), regardless (3.85), support (3.83), weight (3.78), spectrum (3.78), cues (3.74), lights (3.73), situations (3.72), society (3.69), without (3.69), isolating (3.69), raise (3.67), youarenotalone (3.65), difficult (3.63), dedefness (3.62), isolated (3.62), cane (3.62), senses (3.61), public (3.59), diversity (3.59), share (3.59), read (3.53), remember (3.51), loud (3.51), voice (3.49), battle (3.49), acceptance (3.46), questions (3.43), strengths (3.43), deafawareness (3.38), find (3.38), frustration (3.37), someone (3.37), given (3.34), encourage (3.33), advocacy (3.28), due (3.28), misconceptions (3.28), extra (3.25), hearing (3.23), helpful (3.2), strength (3.2), use (3.19), rights (3.19), visually (3.19), familiar (3.18), fog (3.18), therapy (3.18), guide (3.15), focus (3.15), tasks (3.15), accessibilitymatters (3.11), born (3.1), calming (3.1), fullest (3.1), fully (3.09), impaired (3.09), signlanguage (3.09), different (3.07), when (3.06), exhausting (3.06), harder (3.06), lack (3.06), rely (3.06), create (3.03), keep (3.03), victory (3.03), perspective (3.02), makes (3.02), living (3.02), birth (3.01), chest (3.01), supportive (3.01), people (3.0), stereotypes (2.97), break (2.97), monumental (2.95), moment (2.95), mentalhealth (2.94), promote (2.94), with (2.94), culture (2.94), braille (2.93), found (2.92), strategies (2.92), open (2.89), express (2.89), american (2.89), textures (2.89), listening (2.87), sending (2.87), bus (2.86), sight (2.85), sensoryoverload (2.85), overcome (2.84), isnt (2.82), does (2.82), participate (2.82), managed (2.82), noises (2.81), inaccessible (2.81), reaching (2.79), explain (2.79), stigma (2.78), dont (2.78), myday (2.78), impairment (2.76), dressed (2.76), neurological (2.76), determined (2.75), limitations (2.75), come (2.74), order (2.74), trusty (2.72), felt (2.7), faced (2.7), happy (2.7), screen (2.68), member (2.68), able (2.67), raising (2.66), climbing (2.65), designed (2.64), way (2.63), wheelchairuser (2.63), ramp (2.63), out (2.61), face (2.6), certain (2.59), understood (2.58), wheelchairlife (2.58), equitable (2.58), rather (2.56), representation (2.56), smallvictories (2.56), challenging (2.55), visuallyimpaired (2.54), remind (2.54), apps (2.52), crowded (2.52), richness (2.52), greater (2.52), tricky (2.52), struggled (2.52), typical (2.51), available (2.51), touch (2.51), lego (2.49), puzzle (2.49), lipreading (2.49), differentnotless (2.49), deafandproud (2.49), headphones (2.49), loss (2.49), mean (2.48), group (2.48), silence (2.47), being (2.47), it (2.46), embrace (2.45), primarily (2.44), fluorescent (2.44), autismacceptance (2.44), blindandproud (2.44), similar (2.43), manage (2.42), environments (2.42), overwhelming (2.41), tools (2.41), dealing (2.4), gives (2.4), nevergiveup (2.39), stop (2.38), comfortable (2.38), easier (2.38), surrounding (2.37), hold (2.37), comforting (2.37), wasnt (2.35), foster (2.35), itsokaytonotbeokay (2.34), sighted (2.34), audio (2.34), supermarket (2.34), independence (2.32), battles (2.32), while (2.32), rich (2.3), interactions (2.3), neurotypical (2.29), cope (2.29), fine (2.29), beige (2.29), blindnessawareness (2.29), very (2.28), seek (2.27), they (2.27), as (2.26), struggles (2.25), feelings (2.25), talking (2.25), always (2.25), small (2.24), audiobook (2.24), deaflife (2.24), blindcommunity (2.24), accepting (2.24), mentalhealthawareness (2.24), user (2.24), access (2.23), communitysupport (2.22), everydaylife (2.22), resilience (2.21), voices (2.21), full (2.2), experience (2.19), listened (2.19), allows (2.18), misunderstandings (2.18), invisible (2.18), through (2.18), triumphs (2.18), valued (2.17), noise (2.17), accommodations (2.17), bit (2.16), soothing (2.16), family (2.14), grey (2.13), build (2.13), predictable (2.13), captioning (2.13), heavier (2.13), clearly (2.13), too (2.13), sadness (2.12), helped (2.12), isolation (2.11), chair (2.11), do (2.11), capable (2.11), active (2.1), probably (2.1), constant (2.08), lot (2.08), frustrated (2.07), deserves (2.07), disabilitycommunity (2.07), expressions (2.07), audiobooks (2.07), signing (2.07), written (2.06), class (2.06), body (2.06), celebrating (2.06), deserve (2.05), against (2.05), belonging (2.05), gentle (2.04), will (2.04), staring (2.03), granted (2.03), adapt (2.02), however (2.02), mystory (2.01), darkness (2.01), drawing (2.01), coping (2.01), organizations (2.01), arent (2.01), care (2.0), simplest (2.0), insurmountable (2.0), noticed (1.97), hands (1.97), mentalhealthmatters (1.97), going (1.97)
}
\end{spacing}

\subsection{Table \ref{tab:top-words-personas-merged_reverse} not filtered with EmoLex}
\label{full-topwords-personas-merged-reverse}
\begin{spacing}{0.8}
{\footnotesize
 im (12.3), new (11.97), your (10.76), all (10.07), introduction (9.84), pop (9.66), hiking (8.58), eg (8.29), insert (8.2), passionate (8.07), what (7.48), gratitude (7.45), in (7.39), wait (7.22), excited (7.19), sustainable (7.09), whirlwind (7.01), looking (6.98), photography (6.95), helloworld (6.72), hey (6.69), sourdough (6.68), into (6.58), been (6.57), lifelong (6.52), book (6.48), whats (6.39), comments (6.32), media (6.26), officially (6.21), exploring (6.19), related (6.14), learner (6.08), connecting (6.05), up (5.87), likeminded (5.72), status (5.7), diving (5.69), relevant (5.69), recipes (5.69), finally (5.68), news (5.68), chat (5.67), adventures (5.66), newbeginnings (5.65), ideas (5.52), bookworm (5.41), say (5.38), forward (5.38), recommendations (5.34), coffee (5.31), lets (5.27), amazing (5.27), project (5.23), mention (5.16), absolutely (5.16), day (5.15), soaking (5.12), off (5.06), impact (5.06), the (5.05), lurking (5.0), newbie (4.95), explorer (4.9), hike (4.87), you (4.85), positivity (4.85), hashtags (4.84), growth (4.81), cant (4.76), local (4.75), list (4.7), breathtaking (4.7), projects (4.69), sunset (4.67), inspiration (4.66), soul (4.66), good (4.65), todays (4.64), did (4.64), passion (4.6), avid (4.58), todo (4.57), sunrise (4.57), lover (4.55), recipe (4.49), sustainableliving (4.46), plunge (4.46), ai (4.41), caught (4.4), rollercoaster (4.39), nature (4.38), trails (4.34), emotions (4.28), delicious (4.28), park (4.27), sustainability (4.27), enthusiast (4.26), hasnt (4.25), indie (4.24), travel (4.18), inspire (4.15), beautiful (4.14), baking (4.14), newtosocialmedia (4.06), loves (4.04), personal (4.03), refreshing (4.03), morning (4.02), books (4.02), event (4.01), tech (4.0), last (4.0), newtothis (4.0), innovation (3.99), super (3.98), cooking (3.97), over (3.96), feeling (3.96), current (3.96), newhere (3.94), future (3.91), of (3.91), topic (3.9), hustle (3.89), content (3.87), killer (3.84), productive (3.82), monday (3.81), sluggish (3.81), foodie (3.79), bread (3.79), cultures (3.79), starter (3.75), call (3.75), profession (3.74), experimenting (3.72), jog (3.67), work (3.65), data (3.64), old (3.64), parent (3.63), get (3.62), knowledge (3.55), drop (3.55), excitement (3.54), outdoors (3.51), thought (3.5), newly (3.5), tackling (3.5), how (3.49), joining (3.48), reflecting (3.47), ahead (3.47), mindfulness (3.47), jampacked (3.47), jackofalltrades (3.47), started (3.46), cup (3.46), exciting (3.45), stunning (3.44), next (3.42), digital (3.39), memorable (3.39), perfect (3.39), productivity (3.39), hidden (3.38), single (3.38), thinking (3.37), after (3.36), expand (3.35), latest (3.35), seriously (3.34), creativity (3.32), tackled (3.31), least (3.31), here (3.29), ecofriendly (3.28), yoga (3.27), cozy (3.27), bear (3.27), gems (3.26), whole (3.25), chaos (3.23), adventure (3.23), forget (3.22), reflection (3.21), photos (3.2), jump (3.2), food (3.19), let (3.18), mountains (3.14), goal (3.13), climate (3.13), baker (3.1), volunteering (3.1), views (3.09), personalgrowth (3.09), laundry (3.08), coming (3.06), so (3.03), lex (3.02), minted (3.02), today (3.01), kitchen (3.01), goals (3.01), cheers (3.0), hashtag (3.0), burst (3.0), fun (2.98), power (2.98), catch (2.98), afternoon (2.97), lives (2.96), rewarding (2.96), awesome (2.95), properly (2.95), selfintroduction (2.95), deadlines (2.95), traveling (2.94), bustle (2.94), warm (2.94), explore (2.94), quite (2.93), writer (2.9), thrilled (2.89), inspired (2.89), dinner (2.88), artificial (2.88), anxious (2.87), completely (2.86), exhilarating (2.86), lookout (2.86), week (2.85), embracing (2.85), letsconnect (2.85), specific (2.85), heres (2.84), fall (2.84), development (2.84), then (2.83), recently (2.83), figured (2.83), years (2.82), discovering (2.82), pasta (2.81), adventureawaits (2.81), thoughtsoftheday (2.81), trends (2.81), hobbyinterest (2.81), field (2.8), past (2.8), farmers (2.8), autumn (2.79), interestshobbies (2.79), traveler (2.79), vibes (2.78), surprisingly (2.78), dove (2.75), etc (2.74), few (2.74), fresh (2.73), science (2.73), crazy (2.72), goodvibes (2.72), booklover (2.71), suddenly (2.71), this (2.71), ridiculously (2.7), id (2.7), novel (2.69), terrifying (2.68), parenthood (2.65), emotional (2.65), spent (2.64), environmental (2.64), dailyadventures (2.64), wow (2.64), informed (2.64), lifejourney (2.64), crisp (2.6), decided (2.6), uncertainty (2.58), energized (2.58), note (2.58), introductions (2.57), newmember (2.57), contentment (2.57), pure (2.57), late (2.56), dailythoughts (2.56), whipped (2.56), role (2.56), highs (2.55), turned (2.55), fellow (2.55), air (2.54), curious (2.54), wild (2.53), start (2.52), positive (2.52), landscape (2.51), took (2.51), privilege (2.51), ages (2.5), homemade (2.5), engineer (2.49), avocado (2.49), success (2.49), party (2.49), long (2.48), adulting (2.48), ending (2.48), hobbiesinterests (2.48), hit (2.48), weather (2.46), machine (2.45), sipping (2.43), wonders (2.43), market (2.42), newconnections (2.41), bloomer (2.41), updates (2.41), involving (2.41), planet (2.41), footprint (2.41), communitys (2.41), developing (2.41), capturing (2.4), emotionaljourney (2.4), turn (2.4), naturelover (2.4), innovative (2.39), juggling (2.39), dayinthelife (2.39), trying (2.39), events (2.36), expect (2.35), real (2.34), political (2.33), total (2.32), communitylove (2.32), lifes (2.32), career (2.32), carbon (2.32), hiker (2.32), zerowaste (2.32), festival (2.32), currentevents (2.32), industry (2.32), momlife (2.32), humbling (2.32), us (2.32), needed (2.31), horizons (2.31), zones (2.31), currently (2.3), lunch (2.29), accomplishment (2.29), selfdoubt (2.29), highly (2.28), a (2.28), software (2.27), interests (2.27), truly (2.26), intelligence (2.26), rest (2.25), nothing (2.25), photographer (2.24), dream (2.24), aspiring (2.24), heirloom (2.23), eagle (2.23), parenting (2.23), lifelonglearner (2.23), backpacking (2.23), growthmindset (2.23), pondered (2.23), holds (2.23), dive (2.23), date (2.22), guys (2.22), destination (2.22), gorgeous (2.22), developments (2.22), pondering (2.21), adventurer (2.21), apart (2.21), stumbled (2.21), upon (2.21), spots (2.21), sporting (2.21), vegan (2.21), eager (2.21), introduce (2.21), messy (2.21), pretty (2.2), walk (2.2), shifted (2.19), grow (2.18), lifted (2.18), lows (2.18), below (2.18), ourselves (2.17), area (2.17), anyway (2.17), fan (2.16), outdoor (2.15), beats (2.15), an (2.14), resist (2.13), sucker (2.13), mondaymood (2.13), nonprofit (2.13), bands (2.13), algorithms (2.13), heartfelt (2.13), run (2.13), collaborate (2.13), fiction (2.13), muchneeded (2.12), year (2.12), cuisines (2.12), winds (2.12), solution (2.12), colleague (2.12), collaboration (2.12), wherever (2.12), from (2.1), called (2.1), just (2.1), inspiring (2.09), definitely (2.09), struck (2.08), kicked (2.08), unwind (2.07), happening (2.06), learning (2.06), spontaneous (2.06), nights (2.06), kindness (2.05), presentation (2.05), along (2.05), selfcare (2.04), quick (2.04), hooked (2.04), sourdoughfail (2.04), cozyvibes (2.04), luck (2.04), existential (2.04), leo (2.04), ripple (2.04), communityvibes (2.04), dailyactivities (2.04), selfreflection (2.04), designer (2.04), indiemusic (2.04), spreadkindness (2.04), dayinmylife (2.04), green (2.04), effect (2.04), stayathome (2.04), seeker (2.04), wanderlust (2.04), volunteer (2.03), blah (2.03), interconnected (2.03), months (2.03), trail (2.03), fluffy (2.03), such (2.03), curled (2.02), fastpaced (2.02), marketing (2.02), packed (2.02), shelter (2.02), chronic (2.02), openminded (2.02), hobby (2.01), tackle (2.01), shift (2.0), caf (2.0), natural (2.0), scenery (2.0), captivating (2.0), terrible (1.99), staying (1.99), write (1.97), mountain (1.97), wrap (1.96), comment (1.96)
}
\end{spacing}

\newpage
\subsection{Sentiment RQ2}

\begin{figure}[ht!]
    \centering
    \begin{subfigure}[b]{0.32\textwidth} 
        \centering
        \includegraphics[width=\textwidth]{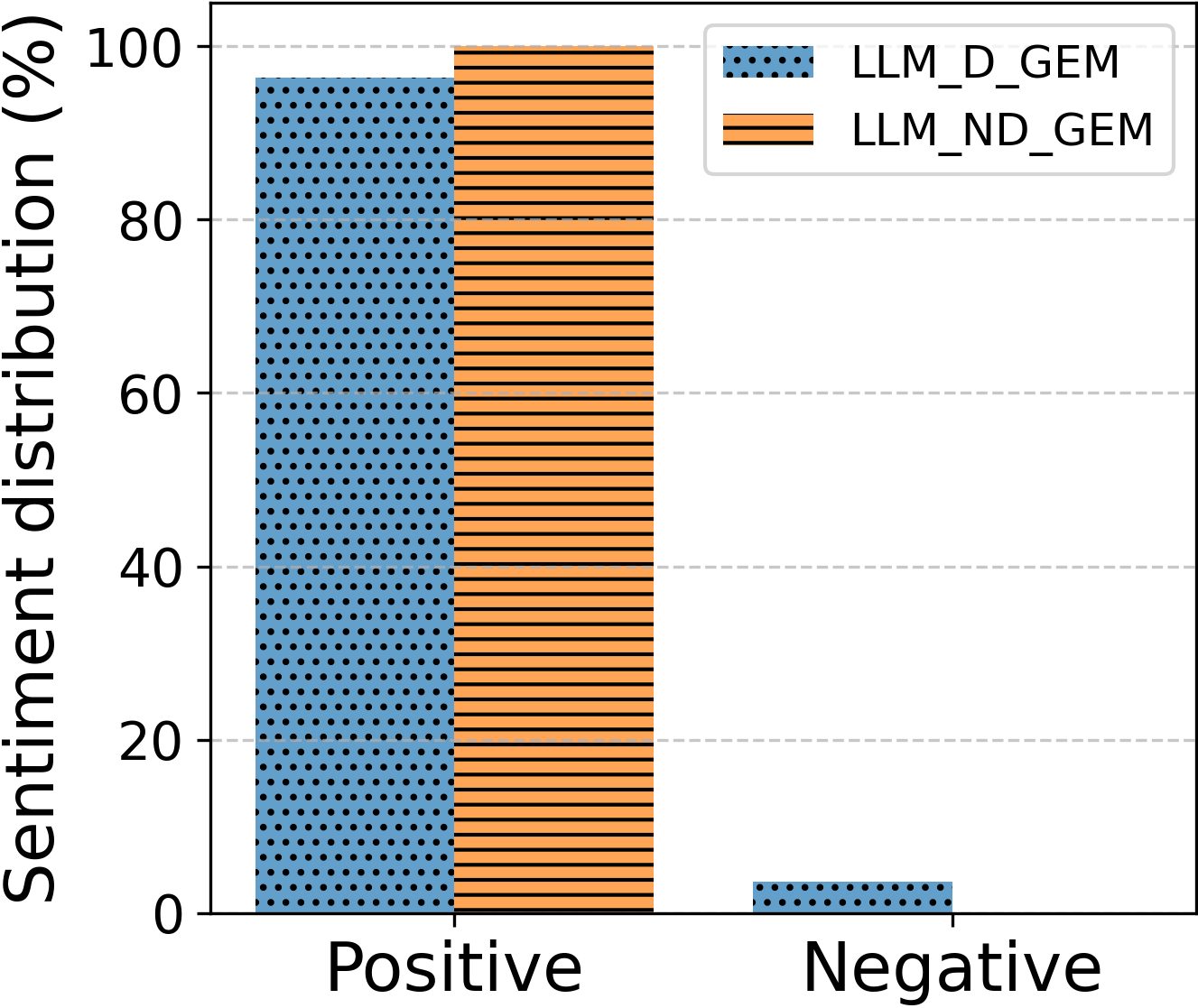}
        \caption{Gemini-1.5F}
                    \Description{Almost all posts of LLM_D_GEM and LLM_ND_GEM have positive sentiment}
        \label{fig:sentiment-persona-gemini}
    \end{subfigure}
    \hfill
    \begin{subfigure}[b]{0.32\textwidth} 
        \centering
        \includegraphics[width=\textwidth]{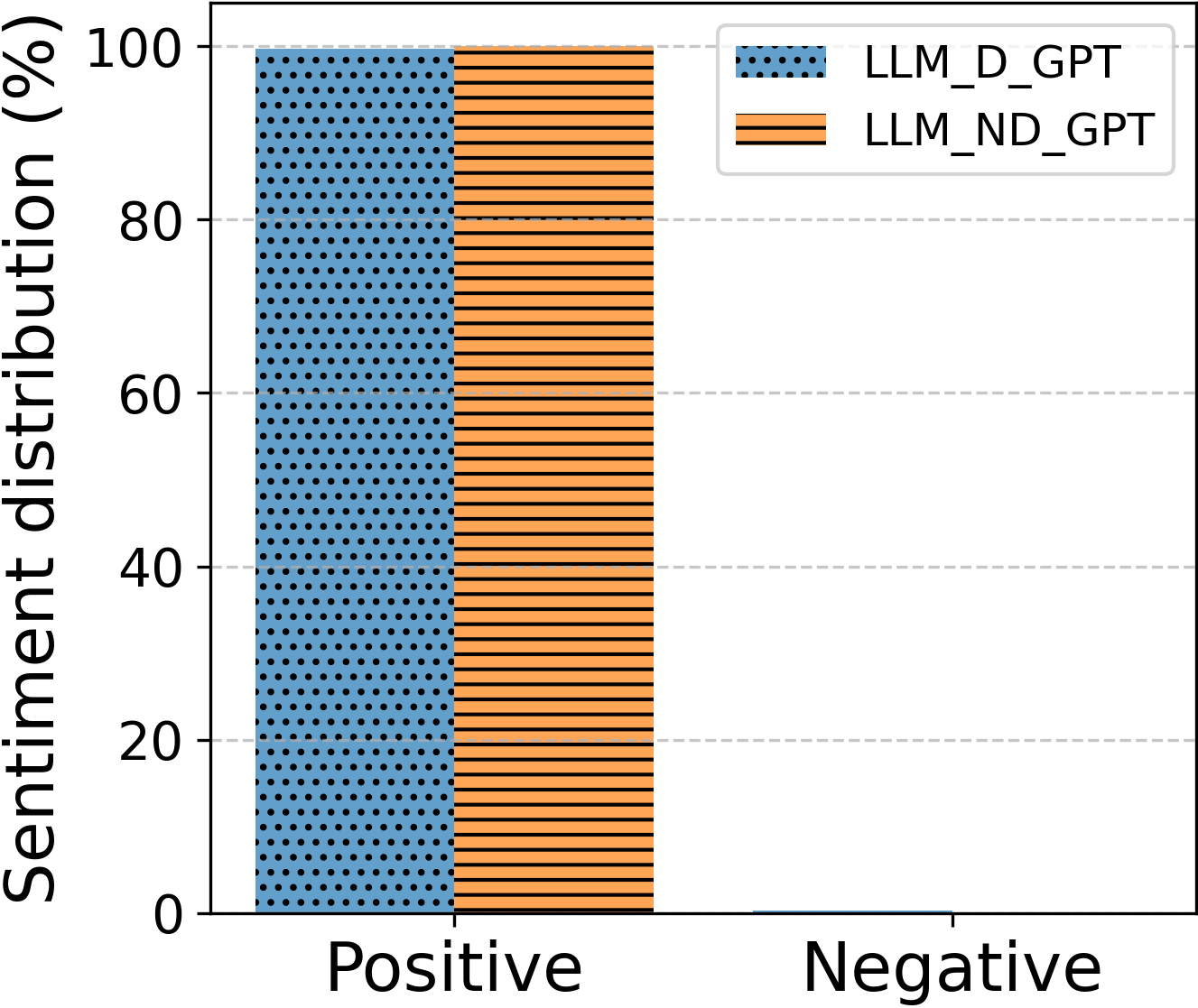}
        \caption{GPT4o-mini}
        \Description{Almost all posts of LLM_D_GPT and LLM_ND_GPT have positive sentiment}
        \label{fig:sentiment-persona-gpt4o}
    \end{subfigure}
    \hfill
    \begin{subfigure}[b]{0.32\textwidth} 
        \centering
        \includegraphics[width=\textwidth]{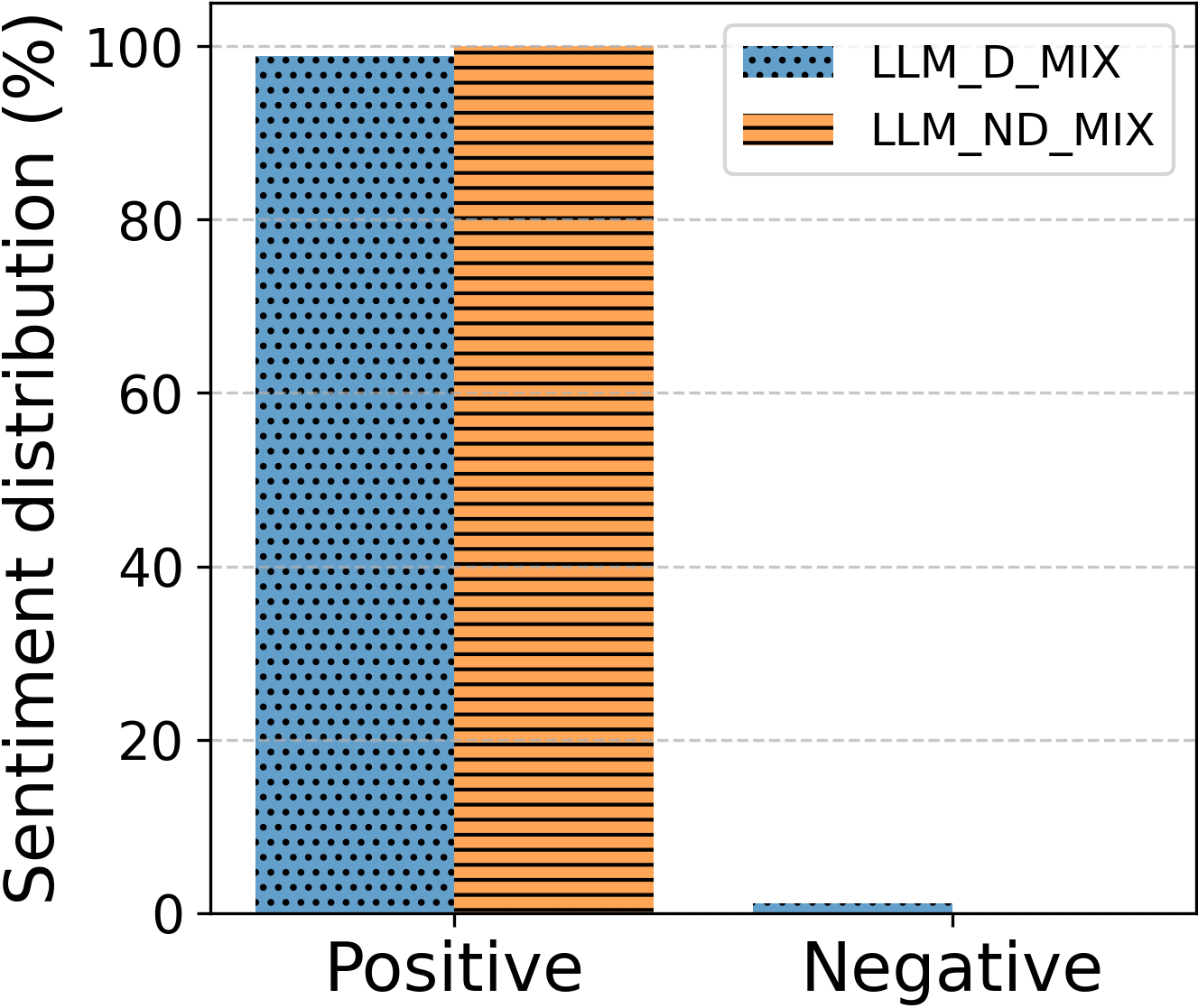}
        \caption{Mixtral-8B}
        \Description{Almost all posts of LLM_D_MIX and LLM_ND_MIX have positive sentiment}
        \label{fig:sentiment-persona-mixtral}
    \end{subfigure}
    
    \caption{Comparison of the sentiment of datasets produced with different LLMs mentioning or not in the prompt that the person has a disability.}
    \label{fig:sentiment-persona-comparison}
\end{figure}

\subsection{Depression RQ2}

\begin{figure}[ht!]
    \centering
    \begin{subfigure}[b]{0.32\textwidth} 
        \centering
        \includegraphics[width=\textwidth]{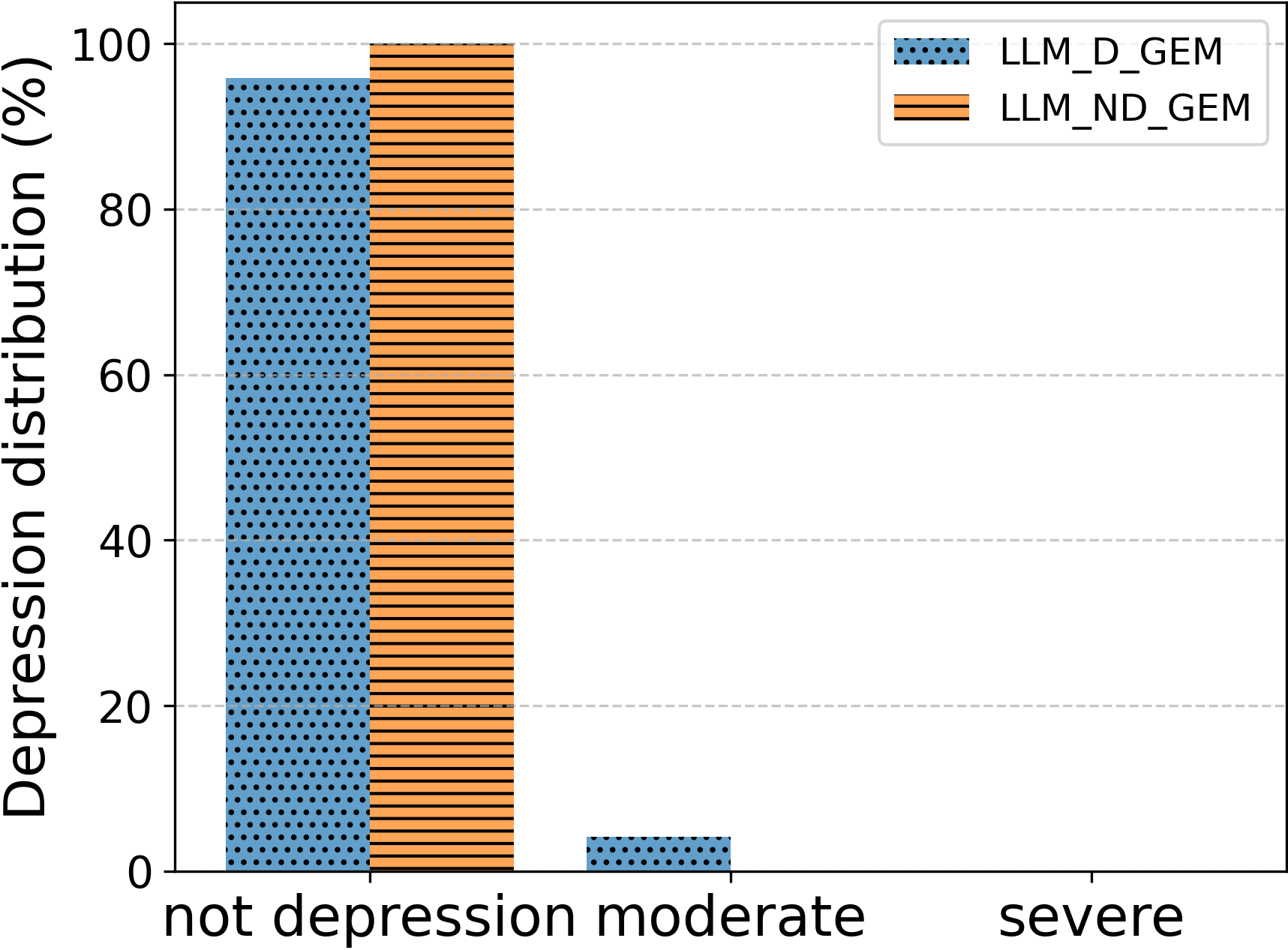}
        \caption{Gemini-1.5F}
                \Description{Almost all posts of LLM_D_GEM and LLM_ND_GEM have not depression}

        \label{fig:depression-persona-gemini}
    \end{subfigure}
    \hfill
    \begin{subfigure}[b]{0.32\textwidth} 
        \centering
        \includegraphics[width=\textwidth]{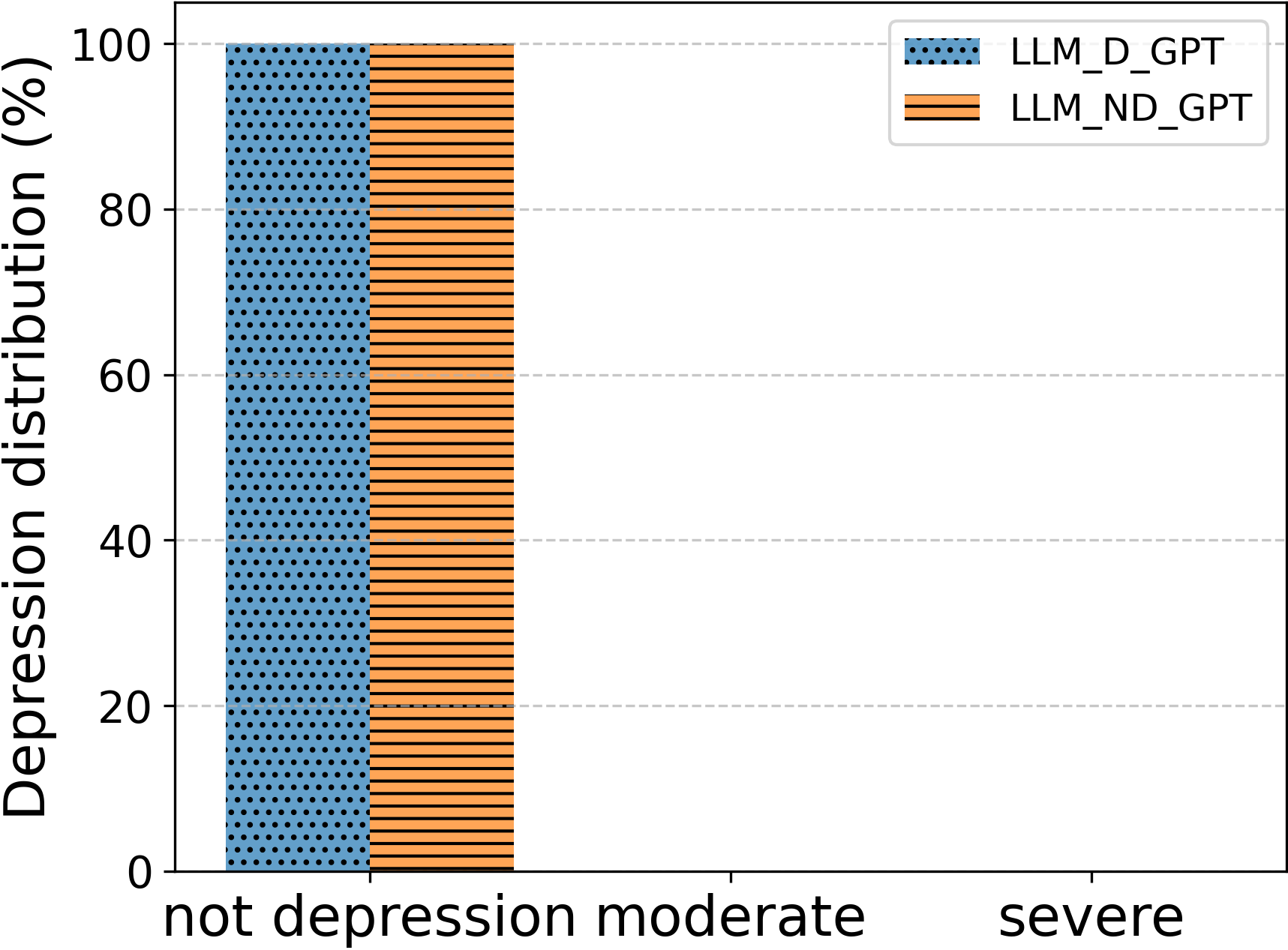}
        \caption{GPT4o-mini}
                        \Description{All posts of LLM_D_GPT and LLM_ND_GPT have not depression}

        \label{fig:depression-persona-gpt4o}
    \end{subfigure}
    \hfill
    \begin{subfigure}[b]{0.32\textwidth} 
        \centering
        \includegraphics[width=\textwidth]{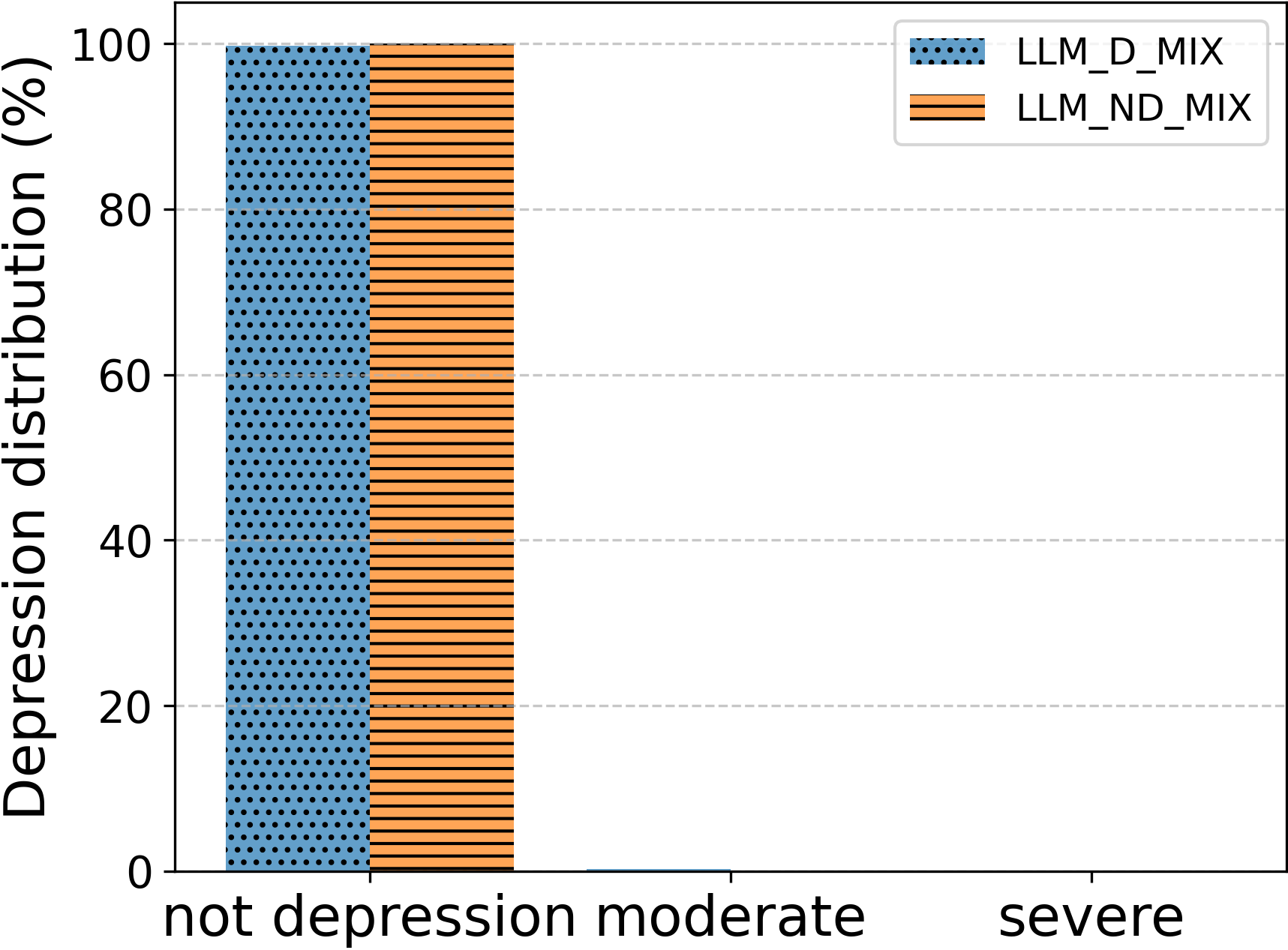}
        \caption{Mixtral-8B}
                        \Description{Almost all posts of LLM_D_MIX and LLM_ND_MIX have not depression}

        \label{fig:depression-persona-mixtral-8b}
    \end{subfigure}
    
    \caption{Comparison of the depression level of datasets produced with different LLMs, mentioning or not in the prompt that the person has a disability.}
    \label{fig:depression-persona-comparison}
\end{figure}

\end{document}